\documentclass[10pt]{article}

\usepackage[utf8]{inputenc}
\usepackage[T1]{fontenc}
\usepackage{times}
\usepackage[margin=1in]{geometry}
\usepackage{hyperref}
\usepackage{url}
\usepackage{amsmath,amsfonts,amssymb,bm,booktabs,multirow,graphicx}
\usepackage{tabularx}
\usepackage{array}
\usepackage{algorithm,algpseudocode}
\usepackage{caption}
\usepackage{subcaption}
\usepackage{xcolor}
\usepackage{enumitem}
\usepackage{natbib}
\setcitestyle{authoryear,round,citesep={;},aysep={,},yysep={;}}

\usepackage{amsmath,amsfonts,bm}

\def\eqref#1{equation~\ref{#1}}

\def\plaineqref#1{\ref{#1}}

\def\1{\bm{1}}

\DeclareMathAlphabet{\mathsfit}{\encodingdefault}{\sfdefault}{m}{sl}
\SetMathAlphabet{\mathsfit}{bold}{\encodingdefault}{\sfdefault}{bx}{n}

\newcommand{\dmdmodel}{DMD-EEG}
\newcommand{\dmdmodelSource}{\dmdmodel{}-Source}
\newcommand{\dmdmodelFull}{\dmdmodel{} (Full)}

\hypersetup{
  colorlinks=true,
  linkcolor=blue!50!black,
  citecolor=green!45!black,
  urlcolor=blue!60!black,
  pdftitle={Separating Diagnosis from Disease Representation: Dual-View EEG Learning with Neural-Dynamics-Guided Deformation}
}

\renewenvironment{abstract}{%
  \small
  \begin{center}%
    {\bfseries \abstractname\vspace{-.5em}\vspace{0pt}}%
  \end{center}%
  \begin{quote}%
}{%
  \end{quote}%
}

\title{Separating Diagnosis from Disease Representation:\\ Dual-View EEG Learning with Neural-Dynamics-Guided Deformation}

\author{%
Jiaying Wang\textsuperscript{1},
Shouqian Shi\textsuperscript{1},
Yutong Chen\textsuperscript{1},
Xu Yang\textsuperscript{1},\\
Jie Chen\textsuperscript{1},
Xingyu Pan\textsuperscript{1},
Lei Zhang\textsuperscript{2},
Sheng Zhong\textsuperscript{1}\\[0.6em]
\textsuperscript{1}Nanjing University \qquad \textsuperscript{2}Tianjin University%
\thanks{Emails: \texttt{wangjiaying@smail.nju.edu.cn}, \texttt{sqlite@nju.edu.cn}, \texttt{hsiaoo@smail.nju.edu.cn}, \texttt{241250115@smail.nju.edu.cn}, \texttt{522026320011@smail.nju.edu.cn}, \texttt{522026320102@smail.nju.edu.cn}, \texttt{lzhang@tju.edu.cn}, \texttt{zhongsheng@nju.edu.cn}.}
}

\date{}

\begin{document}

\sloppy

\maketitle

\begin{abstract}
Electroencephalography (EEG)-based closed-loop neuromodulation calls for a subject-specific structured state, as opposed to a single disease probability, specifying which brain regions are deviant, at which frequencies, and at which lags. Sensor-space models keep the strongest diagnostic evidence without anatomy, source-space models give anatomy at a loss of predictive signal, and post-hoc attributions stay outside the prediction. We separate the two instead of forcing them into one representation, and propose \dmdmodel{} (Dual-view Multiscale Deformation for EEG), which keeps a fixed scalp spectral expert for diagnosis and models the source-space disease-related representation as a low-rank, sparse, iterative deformation of a healthy neural-dynamics prior in a $46$-region-of-interest (ROI) $\times$ $5$-frequency $\times$ $4$-lag (autocorrelation-timescale) space. The two experts meet only at a fixed decision level, so the source state is architecturally separate from the scalp expert. Across major depressive disorder (MDD), first-episode psychosis (FEP), and Parkinson's disease (PD), decision-level fusion matches the strongest single expert on MDD and FEP and exceeds the source branch on PD. On FEP the source expert is the strongest branch, the task where the deformation contributes most. The source state is an explicit ROI--frequency--lag attribution defined in a shared source coordinate system across montages, which we treat as an anatomically-coordinated predictive representation whose coordinates are directly readable and hypothesis-generating. The highest-saliency coordinates align with established disease circuitry (fronto-limbic-temporal regions in MDD, motor-cortex beta in PD), and the MDD state transfers by rank to an unseen cohort recorded with a different montage. Our code is available at \url{https://github.com/zaying/DMD-EEG}.
\end{abstract}

\section{Introduction}

Diagnostic fidelity and representational readability are different quantities. Source reconstruction dilutes scalp spectral discrimination, by $0.08$--$0.17$ area under the ROC curve (AUC) under the strongest power spectral density (PSD) classifier per disease~\citep{welch1967} and by up to $0.16$ AUC under a higher-capacity EEGNet~\citep{lawhern2018eegnet} (Appendix~\ref{app:psd}), yet only the source space carries anatomical meaning. A source-only bottleneck therefore trades away diagnostic signal, and feature concatenation contaminates the source state with sensor-level discriminant information. The objective that maximizes accuracy differs from the objective that yields a readable, anatomically-coordinated disease representation. Clinical and neuromodulation use of EEG needs a structured, real-time state beyond a disease label: where a deviant state occurs, at what frequency, and with what temporal relationship. The central problem is a stable coordinate system for subject-specific neural deviation.

Existing work falls into three groups, none closing the gap. Sensor/source fusion studies fuse at the feature level~\citep{shim2016,huang2026smms}. Without explicit separation constraints, a fused representation leaves the separate contributions of its sensor and source components unidentified~\citep{park2025lmd,locatello2019disentanglement}. Source-space connectivity studies provide ROI-, frequency-, and lagged connectivity, but through fixed estimators or static features~\citep{damborska2020altered,hasanzadeh2020graph,liu2022hypofunction,yang2024timevarying,chen2024dynamicfc,myers2025intracranial,whitton2018eeg}, describing a population-level summary. Post-hoc feature-attribution and visualization methods can reveal a trained predictor's reliance on input features, but discriminative weights in general fall short of identifying the neural sources or mechanisms generating those features~\citep{haufe2014interpretation,schirrmeister2017visualization}. No existing method preserves strong sensor diagnosis, keeps a source representation independent, and makes the disease-related signal an explicit coordinate of the prediction itself.

We address these requirements with two experts combined at a low-capacity decision level: a fixed scalp expert mapping power-spectral evidence to a diagnostic probability, and a source-space expert expressing the disease-related state as an iterative sparse deformation of a subject-conditioned healthy-anchored state $h^0$. The branches share no intermediate learned features or training gradients and interact only at the decision level. With the deformation defined on explicit ROI, frequency, and lag coordinates, the source expert localizes the deviation in terms that can be checked against independent medical literature. We evaluate on three clinical tasks spanning psychiatric and neurological disorders -- major depressive disorder (MDD, MODMA, 24 MDD/29 healthy controls (HC)~\citep{cai2022modma}), first-episode psychosis (FEP, ds003944, 50 FEP/32 HC~\citep{ds003944openneuro}), and Parkinson's disease (PD, ds004584, 100 PD/49 HC~\citep{ds004584openneuro}) -- under a leak-free multi-seed subject-level holdout. These datasets lack subject-level work under a unified protocol. We supply that cross-disease comparison as the benchmark.

Our contributions are:

\begin{itemize}[nosep]
\item \textbf{Dual-view separation of diagnosis and disease representation.} We separate diagnostic fidelity from representational readability in clinical EEG: a fixed low-capacity scalp expert handles diagnosis while a source-space expert learns a disease-related representation in anatomically labeled coordinates, combined only at a fixed decision level. The fusion matches the best single expert within seed noise on MDD and FEP (MDD $0.700$ vs.\ $0.622$; FEP $0.789$ vs.\ $0.794$), halves the source branch's across-seed standard deviation on FEP ($0.139\to0.055$), and exceeds the source branch on PD ($+0.176$ paired $\Delta$AUC).
\item \textbf{Neural-dynamics-guided iterative deformation.} The classification signal is carried by a low-rank iterative sparse deformation of the healthy-anchored state $h^0$ with explicit ROI~$\times$~frequency~$\times$~lag coordinates, so the decision is readable directly on anatomically labeled axes.
\item \textbf{Structured, neurobiologically testable localization.} The decision-sensitivity profile over the healthy-anchored state carries explicit ROI~$\times$~frequency~$\times$~lag coordinates whose highest-saliency entries are directionally consistent with established disease circuitry, and the state is independently classifiable and, for MDD, rank-transferable across acquisition systems. We present this attribution as an anatomically-coordinated, hypothesis-generating predictive representation. It is supported by sensitivity attribution, ROI occlusion, and concordance with the disease literature: MDD on fronto-limbic-temporal regions, PD on motor-cortex beta.
\end{itemize}

\begin{figure*}[t]
\centering
\includegraphics[width=.98\textwidth]{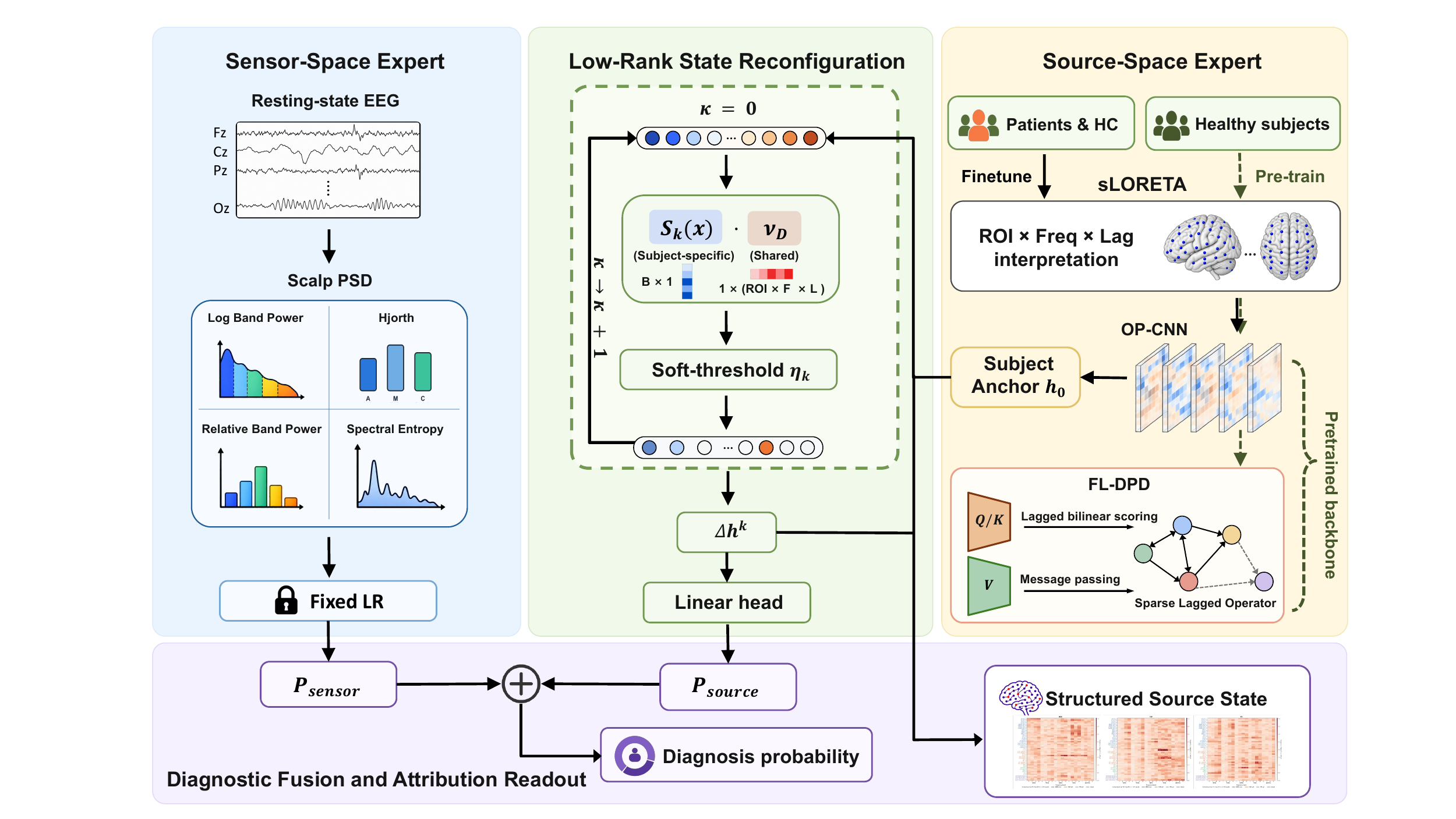}
\caption{\textbf{Dual-view EEG learning.} A fixed PSD--logistic-regression sensor expert maps scalp spectra to a diagnostic probability, and a source expert anchors sLORETA activity to a healthy-dynamics prior (pretrained on healthy EEG only) and deforms it with a low-rank sparse solver into an explicit ROI$\times$frequency$\times$lag state. The two experts combine only at a fixed decision level, leaving the two branches architecturally separate and the source state available for attribution.}
\label{fig:hero}
\end{figure*}

\section{Related Work}

\subsection{Sensor- and source-space EEG for brain disorder recognition}

Sensor-level EEG features have been used for disorder recognition across MDD, FEP, and PD, with recordings, feature definitions, and evaluation protocols varying substantially~\citep{sun2020resting,redwan2024psd,jaramillojimenez2023spectral}. Deep and foundation models on scalp EEG form a parallel representation-learning line~\citep{schirrmeister2017visualization,labram2024,cbramod2025}. Source-space studies add anatomy through reconstruction and connectivity, often with fixed estimators~\citep{damborska2020altered,hasanzadeh2020graph,liu2022hypofunction,yang2024timevarying,chen2024dynamicfc}. Feature-level sensor--source fusion learns a joint representation, but without explicit separation constraints the contribution of each representation stays ambiguous~\citep{park2025lmd,locatello2019disentanglement}. Consequently, a feature-level fusion entangles the source component with the sensor features, leaving the source representation hard to identify on its own. Explicit bottleneck variables keep an intermediate representation that can be inspected and intervened on directly~\citep{koh2020cbm}. Our source state is a learned deformation in anatomical coordinates, distinct from their human-defined concepts. Prior work asks whether combining representations improves classification. We ask how to combine predictive evidence while preserving a source representation for attribution analysis.

\subsection{Directed and frequency-specific brain communication}

Directed and frequency-specific connectivity is well established (Granger causality~\citep{granger1969causal}, directed transfer function~\citep{kaminski1991dtf}, partial directed coherence~\citep{baccala2001pdc}, phase transfer entropy~\citep{lobier2014pte}) and applied to depression and psychosis EEG~\citep{damborska2020altered,hasanzadeh2020graph,liu2022hypofunction,yang2024timevarying,chen2024dynamicfc,kabbara2022eeg}. Intracranial electrophysiology has likewise been used to examine directed prefrontal connectivity in MDD~\citep{myers2025intracranial}. These estimate a graph for classification or group comparison, whereas our source disease representation is a learned deformation state under a healthy-dynamics prior, retained for downstream analysis.

\subsection{Counterfactual, interpretable, and neurobiologically validated modeling}

Post-hoc feature attribution and visualization provide useful model interpretation, but by themselves leave a forward disease representation undefined~\citep{haufe2014interpretation,schirrmeister2017visualization}. Normative modeling frames patient deviation from a healthy distribution~\citep{marquand2016normative,marquand2019conceptual}, and data-driven subtype models decompose heterogeneous disorders into trajectories~\citep{chen2023stratification,young2018subtype}. Source-space EEG normative models predict clinical outcomes from each subject's deviation from a healthy norm~\citep{tong2024individual}. Our deformation is supervised directly by disease labels, which makes it a discriminative state, distinct from an estimated healthy-counterfactual deviation. It carries the classification decision and is defined on explicit neurophysiological coordinates. Each coordinate of the healthy-anchored state has one neural meaning, so it is testable on two grounds: state stability and coordinate readability, and concordance of the highest-saliency coordinates with known disease circuitry. ROI, network, and frequency findings are checked against disease neuroscience, and lag through cross-fold stability, synthetic recovery, and lag perturbation (Appendix~\ref{app:lag}).

\section{Method}

\subsection{Overview}

\label{sec:overview}
The framework combines two independent experts through a low-capacity fusion (Figure~\ref{fig:hero}). The Sensor Expert maps scalp PSD~\citep{welch1967} through a fixed logistic regression to $p_{\mathrm{sensor}}$. The Source Expert source-reconstructs the EEG with a healthy-pretrained oscillatory model and applies a low-rank iterative deformation. A linear head then reads the resulting state to produce $p_{\mathrm{source}}$. The final decision is the fixed average of the two experts,
\begin{equation}
p_{\mathrm{final}}=\tfrac{1}{2}p_{\mathrm{sensor}}+\tfrac{1}{2}p_{\mathrm{source}}.
\label{eq:fusion}
\end{equation}
which keeps the two branches architecturally separate and the scalp expert purely diagnostic. A learnable weight $\alpha$ is evaluated as a secondary ablation (Appendix~\ref{app:fusion}).

\subsection{ROI atlas design}

We use a fixed 46-ROI analysis atlas, chosen before final model evaluation to balance transdiagnostic coverage against the resolution scalp-EEG source reconstruction can support. The ROIs are label-independent and cover MDD, FEP, and PD circuits, so all three disorders share one anatomical coordinate system. Deep structures such as basal ganglia are excluded because routine scalp EEG and sLORETA cannot resolve their subnuclei, so PD interpretation is restricted to cortical, thalamocortical, and large-scale network consequences. The full list, with the design rationale for each region, is in Table~\ref{tab:roi-atlas-complete}.

\subsection{Healthy-dynamics pretraining}

Healthy pretraining never sees a disease label. Scalp EEG is source-reconstructed with sLORETA~\citep{pascualmarqui2002sloreta} on a template head model, and each ROI's time series is then decomposed into five bands ($\delta,\theta,\alpha,\beta,\gamma$). The oscillation-preserving CNN (OP-CNN) is a causal dilated-convolution encoder that preserves ROI, band, and temporal structure. The lag axis uses the per-ROI, per-band autocorrelation function (ACF) of the latent time series at $\tau\in\{1,2,4,8\}$ (scale axis), reduced to one scalar per lag, yielding $h_H\in\mathbb{R}^{46\times5\times4}$ with genuine autocorrelation timescales, free of synthesized coordinates or inter-regional transmission delays. Pretraining optimizes local self-prediction plus cross-ROI communication reconstruction on healthy data only, jointly training the sparse static scaffold $W_H$ and the frequency--lag directed pathway (FL-DPD) generator that produces a communication state
\begin{equation}
A_H=W_H+G_H,\qquad G_H=\tanh\!\big(h_{H,i}+h_{H,j}+b_{f,\ell}\big),
\label{eq:healthy-comm}
\end{equation}
with $G_H$ normalized per sample. Pretraining therefore learns a \emph{healthy neural-dynamics prior}: source EEG maps to an oscillatory local state and then to cross-ROI delayed prediction. The healthy-only corpus pools 269 subjects from seven healthy datasets (Section~\ref{sec:pretrain-data}).

\subsection{Neural-dynamics-conditioned iterative disease deformation}

\label{sec:deform}
The healthy model provides each subject with a subject-conditioned anchor: the band-preserving source-space representation from the oscillation-preserving encoder, with a real per-$\tau$ autocorrelation lag axis, compressed by a small multilayer perceptron (MLP) applied independently per (ROI, band, lag) coordinate into $h^0=h_H(x)\in\mathbb{R}^{920}=\mathbb{R}^{46\times5\times4}$, so each output coordinate keeps its ROI, band, and lag identity. Here ``healthy'' denotes the pretraining objective that shapes the encoder into a healthy-dynamics prior, distinct from a frozen template: each anchor comes from that encoder applied to the subject's own source activity. We write $h^0$ as $h_{\mathrm{health}}$ when we attribute the decision to its coordinates. Disease is then a $K$-step sparse deformation of $h^0$,
\begin{equation}
h^k=h^0+\Delta h^k,\qquad \Delta h^0=0,\qquad k=1,\dots,K.
\label{eq:trajectory}
\end{equation}
The deformation is restricted to a low-rank, cohort-shared disease subspace. A shared direction matrix $V_D\in\mathbb{R}^{r\times 920}$ defines $r$ disease axes, and a small amplitude network predicts a per-subject, per-step signed coefficient $s_k(x)\in\mathbb{R}^{r}$ from the anchor state $h^0(x)$, conditioning each step on the subject's neural dynamics,
\begin{equation}
U_k(x)=V_D^\top s_k(x),
\label{eq:lowrank}
\end{equation}
so all subjects traverse the same shared disease directions with different signed distances, a strong capacity prior for small cohorts. This is an explicit modeling choice, fixing the subject-conditioned quantity as the signed position along a few disease axes shared across the cohort. We refer to this iterative deformation module as the solver. Each solver step applies three operations in sequence. First, the current deformation is moved along the low-rank direction with a per-step step size $\eta_k$:
\begin{equation}
\tilde\Delta h^{k+1}=\Delta h^k-\eta_k\, U_k(x).
\label{eq:gradstep}
\end{equation}
Second, a per-step soft threshold $\tau_k$ shrinks small entries toward zero. Third, a hard top-$m$ selection keeps only the $m$ largest-magnitude coordinates, so at most $m$ of the $920$ entries are nonzero at each step:
\begin{equation}
\Delta h^{k+1}=\operatorname{TopK}_m\!\Big[ \operatorname{SoftThreshold}\!\big( \tilde\Delta h^{k+1},\, \tau_k \big) \Big],
\label{eq:solver}
\end{equation}
where $\operatorname{SoftThreshold}(z,\tau)=\operatorname{sign}(z)\max(|z|-\tau,0)$. The budget $m{=}460$ (top-half of the $920$ coordinates, fixed before training) caps the deformation at $m$ locatable ROI--band--lag coordinates. A fixed fraction is preferable to a single magnitude threshold, which can collapse the support as per-subject magnitudes drift. Retaining the top half prunes the low-magnitude tail while keeping enough candidate coordinates, balancing representational richness against signal purity. Step size and threshold are learned per step and bounded by the solver's parameterization, $\eta_k\in[10^{-4},10^{-1}]$ and $\tau_k\in[0,10^{-3}]$. Because each step adds $-\eta_k U_k$ to all $920$ coordinates before selection, the top-$m$ support is re-estimated at every step: a coordinate suppressed at step $k$ can re-enter at step $k+1$ once its magnitude returns to the top $m$. Each step carries its own amplitude, step size, and threshold, so the $K$ steps jointly refine one sparse deformation and $\Delta h^K$ is a nonlinear function of the per-step signed amplitudes $s_1(x),\dots,s_K(x)$. The decision is read by a light linear head applied directly to $\Delta h^K$, with the solver parameters trained together with the backbone. The deformation $\Delta h^K$ is therefore a source-space discriminative representation learned under disease-label supervision. The full communication pathway $A^K=W_H+G(h^K)$ reconstructs cross-ROI healthy dynamics during pretraining but stays outside the decision readout, since re-fitting it on tens of disease subjects would overfit, and we retain it as a component for larger cohorts. The solver is trained with the minimal objective
\begin{equation}
\mathcal L=\operatorname{CE}\!\big(\hat y^K,y\big)+\lambda\,\operatorname{mean}\!\big(|\Delta h^K|\big),
\label{eq:loss}
\end{equation}
with $\lambda=0.01$ unless stated otherwise. The $\ell_1$ term suppresses deformation entries that fail to earn their place, without forcing $\Delta h^K=0$. The per-step update in Eq.~\plaineqref{eq:solver} never uses the disease label, so inference on unseen subjects is label-free: $x_{\mathrm{unknown}}\to\Delta h^1\to\cdots\to\Delta h^K\to\hat y$. The label is used in the outer objective to train the network, and the trajectory $(\Delta h^1,\dots,\Delta h^K)$ is an output of the forward computation. Because the ROI~$\times$~frequency~$\times$~lag structure of $\Delta h^K$ is a model output, each coordinate carries its own ROI and band semantics. The deformation is therefore comparable across subjects and disorders in one coordinate system. Three quantities answer different questions and are reported as distinct layers: $\Delta h^K$ is the deformation that carries the classification signal, $\partial\,\mathrm{logit}/\partial h_{\mathrm{health}}$ is the decision-sensitivity profile on the anchor from which our attribution is computed, and the Cohen's $d$ of $h^K$ is a descriptive group-difference map.

\subsection{Independent sensor diagnostic expert}

The sensor expert is fixed and simple. For each scalp channel, a power spectral density estimate~\citep{welch1967} on $1$--$45$~Hz segments is summarized into absolute, relative, and log band power per canonical band ($\delta,\theta,\alpha,\beta,\gamma$), plus spectral entropy and the Hjorth activity, mobility, and complexity descriptors~\citep{hjorth1970}, and mapped through a logistic regression to $p_{\mathrm{sensor}}$. This preserves high-fidelity spectral diagnostic evidence without disease-specific classifier selection or contamination of the source latent. Strong cross-validated classifiers are reported separately as baselines.

\section{Experiments}

\subsection{Experimental Setup}

\paragraph{Datasets.}\label{sec:pretrain-data} The backbone is pretrained on 269 healthy subjects across seven datasets~\citep{ds006525openneuro,ding2022brainwide,ds006466openneuro,ds006480openneuro,ds006171openneuro,ds006446openneuro,ds006547openneuro,ds006437openneuro,wu2026hypnotherapy}, with subject-level grouping enforced. These corpora are separate from the three disease datasets, and subject-ID disjointness from the MODMA, ds003944, and ds004584 HCs is verified. Disease adaptation is evaluated on the three tasks of Table~\ref{tab:datasets}. All signals are band-pass filtered ($0.5$--$70$~Hz) with a $50$~Hz notch, re-referenced to the common average, and source-reconstructed over the same 46-ROI atlas.

\paragraph{Evaluation Protocol.} All model results use a leak-free protocol: for each seed $s \in \{42, 99, 2026\}$, a disjoint 20\% subject holdout is locked as the test set, the remaining pool is split into 3 subject-level folds used only to select the fold-best checkpoint, and the inner-best model is evaluated exactly once on that seed's holdout. Holdouts are non-overlapping and excluded from training, early stopping, and selection. The reported value is mean$\pm$std over $n{=}3$ seeds. We report AUC, balanced accuracy (BAcc), area under the precision-recall curve (AUPRC), F1, and cross-seed standard deviation. A fixed five-fold protocol is used only for development ablations (Section~\ref{sec:ablation}) and is noted as selection-biased.

\begin{table}[t]
\caption{Clinical cohorts. All model results use strict subject-level splits.}
\label{tab:datasets}
\centering
\begin{tabular}{lccccc}
\toprule
Disease & Dataset & Patients & HC & Elec. & Rate \\
\midrule
MDD & MODMA~\citep{cai2022modma} & 24 & 29 & 128 & 250 Hz \\
FEP & ds003944~\citep{ds003944openneuro} & 50 & 32 & 60 & 500 Hz \\
PD & ds004584~\citep{ds004584openneuro} & 100 & 49 & 64 & 500 Hz \\
\bottomrule
\end{tabular}
\end{table}

\paragraph{Baselines.}
We reproduced the principal subject-level baselines under the same strict holdout protocol: scalp PSD~\citep{welch1967} with logistic regression, linear support vector machine (SVM)~\citep{cortes1995}, and random forest~\citep{breiman2001}; EEGNet~\citep{lawhern2018eegnet}; DeeperBrain~\citep{wang2026deeperbrain}; CBraMod (official checkpoint); microstate+SVM for FEP~\citep{hill2026microstate}; and spectral random forest for PD~\citep{chen2026gumbel}. Table~\ref{tab:strict-baselines} (Appendix~\ref{app:baseline-repro}) lists all reproduction results with \dmdmodelSource{} and \dmdmodelFull{}. Scalp PSD + LR is also the fixed Sensor Expert inside \dmdmodel{}. Each learning-based baseline used its own preprocessing, scaling, segmentation, and, where available, official code and checkpoint. Published participant-level reports on these datasets~\citep{wang2026deeperbrain,lin2026deptf,hill2026microstate,chen2026gumbel,sslneuro2025} use different protocols, so their numbers resist direct comparison. Every method is reported under one unified subject-level protocol.

\begin{table*}[tb]
\caption{Strict subject-level holdout comparison under a unified local reproduction protocol (three seeds; mean$\pm$std). All learning-based baselines were reproduced under the same subject-level evaluation protocol. Scalp PSD + LR is also used as the fixed Sensor Expert in \dmdmodel{}; \dmdmodelSource{} is the source-space neural-dynamics deformation branch; \dmdmodelFull{} combines the two experts with fixed equal-weight decision fusion ($\alpha{=}0.5$). AUPRC and F1$_{\mathrm{pos}}$ use disease as the positive class. Within each task and metric, \textbf{bold} marks the best result and \underline{underline} the second best.}
\label{tab:strict-baselines}
\centering\small
\resizebox{\textwidth}{!}{%
\begin{tabular}{llccccc}
\toprule
Task & Method & BAcc & AUC & AUPRC & F1$_{\mathrm{pos}}$ \\
\midrule
MDD & Scalp PSD + LR~\citep{welch1967} & \textbf{0.656$\pm$0.106} & 0.622$\pm$0.129 & 0.652$\pm$0.169 & \textbf{0.598$\pm$0.109}\\
MDD & Scalp PSD + RF~\citep{breiman2001} & 0.594$\pm$0.153 & \underline{0.667$\pm$0.109} & \underline{0.716$\pm$0.125} & 0.515$\pm$0.205\\
MDD & Scalp PSD + SVM~\citep{cortes1995} & 0.478$\pm$0.028 & 0.344$\pm$0.042 & 0.463$\pm$0.048 & 0.395$\pm$0.088\\
MDD & EEGNet~\citep{lawhern2018eegnet} & 0.578$\pm$0.058 & 0.522$\pm$0.087 & 0.599$\pm$0.078 & 0.555$\pm$0.060\\
MDD & DeeperBrain~\citep{wang2026deeperbrain} & 0.500$\pm$0.000 & 0.589$\pm$0.016 & 0.608$\pm$0.044 & 0.417$\pm$0.295\\
MDD & CBraMod~\citep{cbramod2025} & 0.483$\pm$0.024 & 0.556$\pm$0.042 & 0.548$\pm$0.021 & 0.550$\pm$0.106\\
\midrule
MDD & \textbf{\dmdmodelSource{}} & 0.622$\pm$0.110 & 0.622$\pm$0.134 & 0.663$\pm$0.140 & 0.548$\pm$0.034\\
MDD & \textbf{\dmdmodelFull{}} & \underline{0.628$\pm$0.136} & \textbf{0.700$\pm$0.109} & \textbf{0.758$\pm$0.110} & \underline{0.583$\pm$0.118}\\
\midrule
FEP & Scalp PSD + LR~\citep{welch1967} & 0.656$\pm$0.077 & 0.700$\pm$0.062 & 0.824$\pm$0.034 & 0.723$\pm$0.069\\
FEP & Scalp PSD + RF~\citep{breiman2001} & 0.517$\pm$0.027 & 0.617$\pm$0.041 & 0.768$\pm$0.071 & 0.665$\pm$0.026\\
FEP & Scalp PSD + SVM~\citep{cortes1995} & 0.500$\pm$0.000 & 0.683$\pm$0.095 & 0.801$\pm$0.058 & \underline{0.769$\pm$0.000}\\
FEP & EEGNet~\citep{lawhern2018eegnet} & 0.611$\pm$0.046 & 0.639$\pm$0.083 & 0.753$\pm$0.047 & 0.761$\pm$0.064\\
FEP & DeeperBrain~\citep{wang2026deeperbrain} & 0.500$\pm$0.000 & 0.639$\pm$0.091 & 0.791$\pm$0.032 & \underline{0.769$\pm$0.000}\\
FEP & CBraMod~\citep{cbramod2025} & 0.511$\pm$0.052 & 0.406$\pm$0.102 & 0.621$\pm$0.066 & 0.699$\pm$0.082\\
FEP & Microstate+SVM~\citep{hill2026microstate} & 0.500$\pm$0.000 & 0.417$\pm$0.130 & 0.625$\pm$0.079 & \underline{0.769$\pm$0.000}\\
\midrule
FEP & \textbf{\dmdmodelSource{}} & \textbf{0.706$\pm$0.131} & \textbf{0.794$\pm$0.139} & \textbf{0.894$\pm$0.081} & \textbf{0.791$\pm$0.031}\\
FEP & \textbf{\dmdmodelFull{}} & \underline{0.689$\pm$0.068} & \underline{0.789$\pm$0.055} & \underline{0.883$\pm$0.036} & 0.765$\pm$0.049\\
\midrule
PD & Scalp PSD + LR~\citep{welch1967} & \textbf{0.700$\pm$0.035} & 0.772$\pm$0.014 & \underline{0.871$\pm$0.022} & 0.784$\pm$0.035\\
PD & Scalp PSD + RF~\citep{breiman2001} & 0.642$\pm$0.094 & 0.718$\pm$0.058 & 0.850$\pm$0.025 & \underline{0.796$\pm$0.064}\\
PD & Scalp PSD + SVM~\citep{cortes1995} & 0.575$\pm$0.074 & \textbf{0.803$\pm$0.051} & \textbf{0.878$\pm$0.028} & \textbf{0.809$\pm$0.024}\\
PD & EEGNet~\citep{lawhern2018eegnet} & 0.658$\pm$0.065 & 0.743$\pm$0.087 & 0.854$\pm$0.046 & 0.600$\pm$0.062\\
PD & DeeperBrain~\citep{wang2026deeperbrain} & 0.592$\pm$0.066 & 0.680$\pm$0.064 & 0.825$\pm$0.040 & 0.762$\pm$0.027\\
PD & CBraMod~\citep{cbramod2025} & 0.642$\pm$0.082 & 0.712$\pm$0.032 & 0.834$\pm$0.008 & 0.611$\pm$0.171\\
PD & Spectral RF~\citep{chen2026gumbel} & 0.542$\pm$0.031 & 0.702$\pm$0.104 & 0.840$\pm$0.067 & 0.774$\pm$0.007\\
\midrule
PD & \textbf{\dmdmodelSource{}} & 0.608$\pm$0.077 & 0.638$\pm$0.020 & 0.767$\pm$0.032 & 0.783$\pm$0.013\\
PD & \textbf{\dmdmodelFull{}} & \underline{0.683$\pm$0.012} & \underline{0.780$\pm$0.004} & 0.864$\pm$0.022 & 0.792$\pm$0.025\\
\bottomrule
\end{tabular}}
\end{table*}

\subsection{Main Results}
\label{sec:mainresults}

\paragraph{Diagnostic fidelity and anatomical structure in the two views.}

\begin{table}[tb]
\caption{\textbf{Complementarity of sensor and source views in \dmdmodel{}.} The Sensor Expert is a fixed Scalp PSD + logistic-regression classifier; the Source Expert predicts directly from the deformation state $\Delta h$ (\dmdmodelSource{}), with solver depth selected per dataset ($K{=}4,r{=}1$ for MDD; $K{=}3,r{=}1$ for FEP; $K{=}2,r{=}2$ for PD). \dmdmodelFull{} fuses the two experts with fixed equal weights ($\alpha{=}0.5$). Sensor--Source Corr. is the Spearman correlation between the pooled out-of-sample predictions of the two experts; F1$_{\mathrm{pos}}$ is the positive-class F1. Within each disease and metric, \textbf{bold} marks the best branch and \underline{underline} the second best.}
\label{tab:dualview}
\centering
\resizebox{\textwidth}{!}{%
\begin{tabular}{llccccc}
\toprule
Disease & Branch & BAcc & AUC & AUPRC & F1$_{\mathrm{pos}}$ & Sensor--Source Corr. \\
\midrule
MDD & Sensor Expert (Scalp PSD + LR) & \textbf{0.656 $\pm$ 0.106} & \underline{0.622 $\pm$ 0.129} & 0.652 $\pm$ 0.169 & \textbf{0.598 $\pm$ 0.109} & \multirow{3}{*}{0.46} \\
MDD & Source Expert & 0.622 $\pm$ 0.110 & \underline{0.622 $\pm$ 0.134} & \underline{0.663 $\pm$ 0.140} & 0.548 $\pm$ 0.034 & \\
MDD & \textbf{\dmdmodelFull{}} & \underline{0.628 $\pm$ 0.136} & \textbf{0.700 $\pm$ 0.109} & \textbf{0.758 $\pm$ 0.110} & \underline{0.583 $\pm$ 0.118} & \\
\midrule
FEP & Sensor Expert (Scalp PSD + LR) & 0.656 $\pm$ 0.077 & 0.700 $\pm$ 0.062 & 0.824 $\pm$ 0.034 & 0.723 $\pm$ 0.069 & \multirow{3}{*}{0.25} \\
FEP & Source Expert & \textbf{0.706 $\pm$ 0.131} & \textbf{0.794 $\pm$ 0.139} & \textbf{0.894 $\pm$ 0.081} & \textbf{0.791 $\pm$ 0.031} & \\
FEP & \textbf{\dmdmodelFull{}} & \underline{0.689 $\pm$ 0.068} & \underline{0.789 $\pm$ 0.055} & \underline{0.883 $\pm$ 0.036} & \underline{0.765 $\pm$ 0.049} & \\
\midrule
PD & Sensor Expert (Scalp PSD + LR) & \textbf{0.700 $\pm$ 0.035} & \underline{0.772 $\pm$ 0.014} & \textbf{0.871 $\pm$ 0.022} & \underline{0.784 $\pm$ 0.035} & \multirow{3}{*}{0.17} \\
PD & Source Expert & 0.608 $\pm$ 0.077 & 0.638 $\pm$ 0.020 & 0.767 $\pm$ 0.032 & 0.783 $\pm$ 0.013 & \\
PD & \textbf{\dmdmodelFull{}} & \underline{0.683 $\pm$ 0.012} & \textbf{0.780 $\pm$ 0.004} & \underline{0.864 $\pm$ 0.022} & \textbf{0.792 $\pm$ 0.025} & \\
\bottomrule
\end{tabular}}
\end{table}

With the same PSD classifier applied to both representations, sLORETA source reconstruction dilutes scalp spectral discrimination by $0.08$ to $0.17$ AUC (source-minus-scalp penalty $-0.08$ to $-0.17$; Appendix Table~\ref{tab:psd}): the source space gives anatomical coordinates, but the source-transformation pipeline (inverse reconstruction, regularization, and ROI aggregation) discards part of the sensor-level spectral signal. A stronger classifier recovers only part of this loss: EEGNet~\citep{lawhern2018eegnet} applied to both representations under the identical holdout brings the source representation within $0.02$ AUC of its scalp counterpart on MDD and FEP and $0.16$ AUC below on PD (Table~\ref{tab:psd}). Table~\ref{tab:dualview} reports the dual-view results: the source branch is the strongest single expert on FEP (AUC $0.794$), and the fixed fusion matches the best single branch within seed noise on MDD and FEP and exceeds the source branch on PD (MDD $0.700$, FEP $0.789$, PD $0.780$), with a significant gain over the source branch on PD and a positive trend on MDD, and a preserved strong FEP source branch at half its across-seed standard deviation ($0.139\to0.055$).

\paragraph{Sensor and source experts capture complementary evidence.}
The two views cover different errors, which is what the fusion exploits: their pooled out-of-sample predictions are weakly to moderately correlated (Spearman $r{=}0.46$ MDD, $0.25$ FEP, $0.17$ PD), so they are non-redundant. A paired subject-level bootstrap (Table~\ref{tab:paired-dauc}) gives fusion-minus-source AUC of $+0.067$ MDD, $+0.030$ FEP, $+0.176$ PD, and fusion-minus-sensor of $+0.052$, $+0.065$, $-0.006$. The PD fusion-minus-source gain has a 95\% CI that excludes zero. On MDD the gains are positive trends, and on FEP the fusion matches the strong source branch while halving its across-seed standard deviation. Appendix ROC curves (Figure~\ref{fig:auroc}) and the confidence scatter (Figure~\ref{fig:scatter}) show the same pattern at the subject level. On PD the signal is largely spectral: the scalp classifier is the strongest of any task-and-view combination (AUC $0.803$), both source probes fall well below it (source-ROI PSD $0.640$, source EEGNet $0.587$), consistent with cortical power structure that source reconstruction dilutes plus deep basal-ganglia pathology sLORETA cannot resolve. The source view is thus complementary there, and PD is where the fusion gain over the weaker branch is significant ($\Delta$AUC $+0.18$).

\paragraph{The deformation is an independently predictive, direction-reproducible discriminative axis.}
The source state is independently classifiable (MDD AUC $0.622$, FEP $0.794$, PD $0.638$; Table~\ref{tab:dualview}), and its readout geometry is reproducible: the linear readout direction is cross-seed stable for MDD (signed cosine $0.971$, permutation $p{<}0.001$) and FEP ($0.840$, $p{<}0.001$) in $2/3$ seed pairs, so the model sorts subjects along one consistent axis and every coordinate of the healthy-anchored state maps the decision to a specific ROI, band, and lag. Because single-fold estimates are noisy at this cohort size, we report the consolidated saliency (coordinates direction-consistent across the $3$ splits $\times$ $3$ folds; Table~\ref{tab:attribution}, Figure~\ref{fig:attrib}). The literature prior is fixed in the appendix and never used for training, checkpoint selection, Top-K selection, or hyperparameter tuning.

\subsection{Analysis}

\paragraph{Attribution readout and medical concordance.}
We distinguish three quantities. The anchor $h^0$ is the subject-conditioned, input-defined state whose coordinates carry a stable per-subject anatomical identity, and it is the object the attribution is computed on. $\Delta h^K$ is the solver's deformation carrying the classification signal, and the final state $h^K=h^0+\Delta h^K$ defines the descriptive group-effect map. The attribution below is the decision-sensitivity profile $\partial\,\mathrm{logit}/\partial h_{\mathrm{health}}$, which identifies the anchored coordinates the prediction depends on. Sensitivity and group difference measure different properties: sensitivity measures how much the decision depends on a coordinate, while the group difference measures how much the state varies between groups. A decision-critical coordinate can show little group difference, and a group-different coordinate can leave the decision unchanged, so we report the two separately. High-sensitivity ROIs are hypothesis-generating and prioritize plausible regions for future stimulation-target validation.
We read the decision sensitivity from the healthy-anchored state by the unsigned amplitude saliency $\mathrm{mean}_{\mathrm{subj}}|\partial\,\mathrm{logit}/\partial h_{\mathrm{health}}|$, which is cross-seed stable (axis cosine $0.83$--$0.87$) and agrees with the occlusion test (ROI Spearman $+0.31$ MDD, $+0.59$ PD; Appendix~\ref{app:occlusion}). The signed saliency supplies the band and lag direction (Table~\ref{tab:attribution}, Figures~\ref{fig:attrib},~\ref{fig:glassbrain}). The highest-importance ROIs concentrate on disease-relevant circuitry. For MDD, the prefrontal, temporal, limbic, and default-mode coordinates (DLPFC, MTG, amygdala, parahippocampal cortex, dACC, sgACC, dmPFC, PCC) are broadly concordant with reported large-scale network dysconnectivity in depression~\citep{kaiser2015network,damborska2020altered}. Independent structural work also reports lower hippocampal volume in MDD~\citep{sheline1996hippocampal,videbech2004hippocampal,schmaal2016subcortical}. For PD, the sensorimotor coordinates (M1, premotor, S1, SMA) and beta-weighted profile are qualitatively concordant with reported abnormal motor-cortical beta--gamma coupling and cortico-subthalamic beta connectivity~\citep{dehemptinne2013pac,litvak2011connectivity}. They fall short of establishing elevated beta power in every individual source ROI. For FEP, the fronto-parietal and insular coordinates are consistent with meta-analytic evidence of dysconnectivity across default-mode, salience, and central-executive networks in early psychosis~\citep{oneill2019dysconnectivity}. As an interventional check, zeroing a top-ranked ROI's band--lag block lowers the decision AUC for FEP (SPL, dmPFC, insula), PD (premotor, M1, dACC), and MDD ($dACC_R$; a joint ROI set lowers it further, Appendix~\ref{app:occlusion}), so the decision relies on anatomically plausible coordinates (the atlas is fixed before evaluation, and the literature played no role in training or selection).

\paragraph{External generalization.}
As an external check the trained MDD source branch transfers, in the shared 46-ROI space, to an independent cohort with a different montage (ds003478~\citep{ds003478openneuro}, 86 subjects: 11 current-MDD, 75 low-BDI controls): the ranking AUC is $0.724\pm0.042$ with MDD scoring above controls across seeds, though a fixed $0.5$ threshold misses the ordering, so the transfer evidence is confined to ranking (Table~\ref{tab:external}).

\subsection{Ablation Study}
\label{sec:ablation}

The ablations are controlled single-variable diagnostics (Appendix~\ref{app:dev-ablation}). The one-shot ($K{=}1$) and no-deformation ($K{=}0$) controls stay below the selected depth on every task (Table~\ref{tab:ablation}). The no-deformation control feeds the anchor $h^0$ to the same linear head the solver path uses, isolating the deformation. On FEP the selected solver ($K{=}3$) reaches holdout AUC $0.794$ against $0.678$ at $K{=}0$, and MDD inner-validation AUC drops from $0.801$ to $0.780$, so the deformation carries discriminative signal beyond the anchor it is built from. Removing the healthy pretraining degrades the source branch on every task, most sharply on PD (strict-holdout AUC $0.638\to0.415$; Table~\ref{tab:components}), so the healthy-dynamics pretraining helps the source branch. Removing the multiscale temporal context degrades MDD ranking with BAcc nearly unchanged (Appendix~\ref{app:dev-ablation}). Solver depth and rank are selected per dataset by inner validation ($K{=}4,r{=}1$ MDD; $K{=}3,r{=}1$ FEP; $K{=}2,r{=}2$ PD; Table~\ref{tab:ablation}).

\section{Discussion}

\subsection{The value of separation}

Diagnosis and disease representation have different requirements: a scalp spectral expert preserves the diagnostic evidence that source reconstruction dilutes (Table~\ref{tab:psd}), while a source-space expert produces an anatomically interpretable state, and the two combine at a fixed decision level. The fusion matches the best single branch within seed noise on MDD and FEP, raising MDD AUC from $0.622$ to $0.700$ and preserving the strong FEP source branch at roughly half its variance. On PD it exceeds the source branch, from $0.638$ to $0.780$. Sensor--source correlation stays low as the source branch strengthens (pooled Spearman $0.28\to0.25$ on FEP), indicating complementary predictions. The anatomically-coordinated sensitivity profile is the cross-seed reproducible one.

The low-rank subspace is our modeling choice. The deformation carries classification and targets explicit ROI~$\times$~frequency~$\times$~lag coordinates, so its decision is attributable to anatomically labeled axes. The attribution claim rests on a cross-seed reproducible sensitivity axis (MDD, FEP) and highest-importance coordinates (cross-seed stable for all three tasks) aligned with the implicated circuitry, and we treat it as hypothesis-generating. Read this way, the representation returns a short list of ROI--band--lag coordinates that constrain where mechanistic hypotheses can be tested. This paper solves the estimation side of the closed-loop problem: $EEG_t \to p_{\mathrm{disease},t} + \Delta h_t$, leaving the stimulation-response model $\Delta h_{t+\Delta}=F(\Delta h_t,u_t)$ to future TMS-EEG work (transcranial magnetic stimulation with concurrent EEG). Our contribution is structured state estimation.

\subsection{Limitations}

The clinical cohorts are small and sLORETA constrains deep structures, so the anatomical claims are mesoscale by design: the 46 ROIs are shared source coordinates, adjacent-ROI differences fall short of millimeter-level localization, and basal-ganglia subnuclei are excluded. Lag coordinates are autocorrelation lags with a magnitude-only reading, and the communication readout is a model-implied predictive quantity, stopping short of synaptic conduction delays or a measured causal pathway. Coordinate-level localization relies on the internal stability tests (cross-fold/cross-seed stability, synthetic recovery, and lag perturbation; Appendix~\ref{app:lag}) and on external medical-consistency checks. The fusion uses a fixed $\alpha{=}0.5$, and it matches the best single branch within seed noise, with its advantage concentrated on the weaker branch. Attributions in Table~\ref{tab:attribution} are directional decision-sensitivity findings, and the exact coordinate map is a directional result, with a validated signature left to larger cohorts.

\section{Conclusion}

We separate diagnosis from an anatomically-coordinated disease representation in clinical EEG: a simple scalp expert preserves diagnostic evidence, while a healthy-dynamics-conditioned low-rank deformation model learns an independently predictive source-space state in explicit ROI~$\times$~frequency~$\times$~lag coordinates. Under a strict subject-level protocol, decision-level fusion matches the best branch on MDD and FEP within seed noise and exceeds the source branch on PD. The source branch is the strongest single expert on FEP and direction-reproducible on MDD and FEP. Its ROI--frequency--lag attribution is concordant with known disease circuitry and, for MDD, transfers to a cohort recorded with a different montage. Because this reproducible, anatomically-indexed axis is explicit in anatomical coordinates, it is a structured state that future neuromodulation work could build on.

\section*{Acknowledgments}

We thank the MODMA project team and the OpenNeuro contributors of ds003944, ds004584, and the healthy pretraining corpora for making their data publicly available.

\bibliography{DMD-EEG_arxiv}
\bibliographystyle{DMD-EEG_arxiv}

\appendix
\section{Evaluation protocol details}

\subsection{Leak-free multi-seed holdout}

For each seed $s \in \{42, 99, 2026\}$: a stratified, disjoint 20\% subject holdout is locked as the test set (the three seeds' holdouts are disjoint). The remaining pool is split into 3 subject-level stratified fold pairs (train 2/3, val 1/3). Each fold trains with early stopping and stores its fold-best checkpoint by a subject-level metric. The inner-best checkpoint is selected from the three fold-best checkpoints by inner-validation subject AUC (ties broken by the selection metric), and that one model is evaluated once on the seed's holdout. The reported value is mean$\pm$std across the $3$ seeds ($n{=}3$ independent holdout evaluations); inner-fold scores serve only as selection references. Holdout subjects are excluded from every fold's training and validation, and subject-level clip grouping guarantees no clip of a test subject appears in training.

We use non-overlapping subject partitions by design and report the variance. Statistical significance is assessed via a subject-level bootstrap, independent of the number of seeds.

\subsection{Model selection metric}

Two selection stages use different metrics. Early stopping and the fold-best checkpoint use a subject-level metric, by default the subject-level negative log-likelihood (clip probabilities aggregated per subject). This stabilizes MDD across seeds but is sometimes suboptimal (FEP can prefer clip-level selection), so it is treated as a per-disease setting. The inner-best checkpoint across the three folds, and the solver depth and rank in the $K$/$r$ scan, are then selected by inner-validation subject AUC.

\subsection{Trajectory reporting rules}

A layer $k^*$ is eligible for discussion only if the trajectory type is stable-accumulating or final-layer-selective. Oscillatory/reversing trajectories are marked ``unstable\_oscillatory'' and are withheld from biological interpretation. For every layer we report direction similarity between adjacent $\Delta h$ steps, per-layer $|\Delta h|$, saturation rate, and whether candidate coordinates flip sign across $k=1..K$.

\subsection{Deformation solver as an algorithm}

Algorithm~\ref{alg:solver} summarizes the iterative deformation of Section~\ref{sec:deform}: the per-step update depends only on the source representation, the anchor $h^0$, and the current deformation, with the disease label entering only the outer objective Eq.~\plaineqref{eq:loss}.

\begin{algorithm}[t]
\caption{Neural-dynamics-conditioned iterative disease deformation}
\label{alg:solver}
\begin{algorithmic}[1]
\Require Source representation, healthy model $f_{\theta_H}$, shared directions $V_D$, amplitude network $s$, steps $K$, top-$m$
\Ensure Deformation state $\Delta h^K$, diagnosis $\hat y^K$
\State $h^0 \gets f_{\theta_H}(x)$ \Comment{self-dynamics anchor}
\State $\Delta h^0 \gets 0$
\For{$k=0,\dots,K-1$}
  \State $U_k(x) \gets V_D^\top s_k(x)$ \Comment{low-rank disease direction, per-subject signed amplitude}
  \State $\eta_k \gets \eta_{\min}+(\eta_{\max}-\eta_{\min})\,\operatorname{sigmoid}(\tilde\eta_k);\ \tau_k \gets \tau_{\min}+(\tau_{\max}-\tau_{\min})\,\operatorname{sigmoid}(\tilde\tau_k)$
  \State $\tilde\Delta h^{k+1} \gets \Delta h^k-\eta_k U_k(x)$
  \State $\Delta h^{k+1} \gets \operatorname{TopK}_m\!\big[\operatorname{SoftThreshold}(\tilde\Delta h^{k+1}, \tau_k)\big]$
  \State $\hat y^k \gets \mathrm{Head}_D(\Delta h^{k+1})$ \Comment{delta\_h readout}
\EndFor
\State \Return $\Delta h^K,\ \hat y^K$
\end{algorithmic}
\end{algorithm}

\section{Preprocessing and leakage control}

\subsection{Signal preprocessing}

All signals are band-pass filtered ($0.5$--$70$~Hz) with a $50$~Hz notch, re-referenced to the common average, and source-reconstructed with sLORETA on a shared template head model into the 46-ROI atlas. For the seven healthy pretraining corpora (Section~\ref{sec:pretrain-data}; 269 subjects), to bound the compute budget and balance subjects across cohorts, each subject contributes up to 20 randomly sampled 10\,s segments (50\% overlap), with subject-level grouping enforced. Preprocessing parameters are fixed per dataset family before any split is drawn, so no test-subject statistics enter preprocessing.

\subsection{Leakage control in cross-ROI analyses}

Whenever cross-ROI predictive increments or direction asymmetries are reported, we apply symmetric orthogonalization of the ROI time series to remove zero-lag common-driver components before recomputing the quantity, and we report both raw and leakage-corrected values (see the supplementary source-side analyses). This control is decisive: raw cross-ROI increments disappear after correction, so no claim of directed communication is made from increments alone.

\section{Baseline reproduction details}\label{app:baseline-repro}

All learning-based baselines share the proposed model's subject-level protocol: the same non-overlapping multi-seed holdout, subject-level metric aggregation (clip probabilities averaged per subject), and the restriction that holdout subjects never enter training, early stopping, or model selection. EEGNet is trained from random initialization on the four-band filtered series at the native montage following its published architecture, and its source variant replaces the input with the $46$-ROI source series; CBraMod uses the authors' official checkpoint and fine-tuning pipeline; DeeperBrain follows its publication's protocol, as its training code was unavailable, and its near-chance subject-level results reflect the difficulty of subject-wise generalization on these small cohorts. On MDD and FEP several learning-based baselines collapse to a single predicted class, giving a balanced accuracy of exactly $0.5$ with zero variance; their AUC stays above $0.5$, so this is a threshold degeneracy at these cohort sizes, distinct from a numerical failure. The microstate baseline follows the reported FEP microstate construction, and the PD spectral random forest follows~\citep{chen2026gumbel}. Reproduction choices (input scaling, segmentation length, montage alignment, checkpoint provenance) are fixed per baseline and listed with the shared settings in the result records.

\section{Scalp versus source spectral discrimination}

\label{app:psd}

Table~\ref{tab:psd} applies two probes to the scalp and to the $46$-ROI source representation: a per-disease PSD probe (the best of logistic regression, random forest, and SVM) and EEGNet~\citep{lawhern2018eegnet}. The penalty is source minus scalp within the same probe (negative $=$ weaker source). Comparing the probes tests whether a higher-capacity source classifier closes the gap.

\begin{table}[h]
\caption{Representation dilemma under the strict protocol (AUC). Each method is applied to both representations: the PSD probe uses, for each disease, the best of logistic regression, random forest, and SVM on the PSD features, applied identically to the scalp and source-ROI PSD, and EEGNet~\citep{lawhern2018eegnet} is trained from random initialization on the scalp channels and on the $46$-ROI source time series under the identical holdout. The penalty is source minus scalp within the same method; a negative penalty means the source representation is weaker. Values are subject-level mean AUC from the same runs as Table~\ref{tab:strict-baselines}.}
\label{tab:psd}
\centering
\small
\begin{tabular}{llccc}
\toprule
Disease & Method & Scalp & Source & Penalty \\
\midrule
MDD & PSD & 0.667 & 0.589 & $-0.078$ \\
MDD & EEGNet & 0.522 & 0.533 & $+0.011$ \\
\midrule
FEP & PSD & 0.700 & 0.533 & $-0.167$ \\
FEP & EEGNet & 0.639 & 0.628 & $-0.011$ \\
\midrule
PD & PSD & 0.803 & 0.640 & $-0.163$ \\
PD & EEGNet & 0.743 & 0.587 & $-0.156$ \\
\bottomrule
\end{tabular}
\end{table}

\section{ROI atlas and hyperparameters}

\subsection{ROI atlas (46)}

The atlas is fixed before disease-model evaluation. Each left/right entry below is one separate coordinate, giving $23\times2=46$ ROIs.

\begin{table*}[t]
\caption{Complete fixed 46-ROI atlas. L and R are distinct source coordinates. The names are the analysis labels used throughout the code and result records. Design groups record the inclusion rationale fixed before disease-model evaluation.}
\label{tab:roi-atlas-complete}
\centering\small
\resizebox{\textwidth}{!}{%
\begin{tabular}{rlll}
\toprule
Pair & ROI labels (two coordinates) & Design group & Anatomical name\\
\midrule
1 & DLPFC\_L, DLPFC\_R & MDD core & Dorsolateral prefrontal cortex\\
2 & dmPFC\_L, dmPFC\_R & MDD core & Dorsomedial prefrontal cortex\\
3 & lateral\_OFC\_L, lateral\_OFC\_R & MDD core & Lateral orbitofrontal cortex\\
4 & medial\_OFC\_L, medial\_OFC\_R & MDD core & Medial orbitofrontal cortex\\
5 & dACC\_L, dACC\_R & MDD core & Dorsal anterior cingulate cortex\\
6 & sgACC\_L, sgACC\_R & MDD core & Subgenual anterior cingulate cortex\\
7 & PCC\_L, PCC\_R & MDD core & Posterior cingulate cortex\\
8 & insula\_L, insula\_R & MDD core & Insular cortex\\
9 & amygdala\_L, amygdala\_R & MDD core & Amygdala\\
10 & hippocampus\_L, hippocampus\_R & MDD core & Hippocampus\\
11 & thalamus\_L, thalamus\_R & MDD core & Thalamus\\
12 & IFG\_L, IFG\_R & FEP language/attention & Inferior frontal gyrus\\
13 & STG\_auditory\_L, STG\_auditory\_R & FEP language/attention & Auditory superior temporal gyrus\\
14 & MTG\_L, MTG\_R & FEP language/attention & Middle temporal gyrus\\
15 & precuneus\_L, precuneus\_R & FEP language/attention & Precuneus\\
16 & SPL\_L, SPL\_R & FEP language/attention & Superior parietal lobule\\
17 & IPL\_angular\_L, IPL\_angular\_R & FEP language/attention & Angular part of inferior parietal lobule\\
18 & M1\_L, M1\_R & PD motor & Primary motor cortex\\
19 & premotor\_L, premotor\_R & PD motor & Premotor cortex\\
20 & SMA\_L, SMA\_R & PD motor & Supplementary motor area\\
21 & S1\_L, S1\_R & PD motor & Primary somatosensory cortex\\
22 & parahippocampal\_L, parahippocampal\_R & Robustness extension & Parahippocampal cortex\\
23 & visual\_association\_L, visual\_association\_R & Robustness extension & Visual association cortex\\
\bottomrule
\end{tabular}%
}
\end{table*}

The 22 MDD-core coordinates were expanded before final evaluation by 12 FEP language/attention, 8 PD motor, and 4 robustness coordinates. Basal-ganglia subnuclei are excluded as primary ROIs: template-head sLORETA with routine scalp EEG falls short of supporting independent subnuclear claims.

\subsection{Hyperparameters}

\begin{table}[h]
\caption{Default hyperparameters.}
\label{tab:hyper}
\centering
\begin{tabularx}{\columnwidth}{@{}l X@{}}
\toprule
Parameter & Value \\
\midrule
ROI count & 46 \\
Bands & $\delta,\theta,\alpha,\beta,\gamma$ \\
Lag indices & 4 (real per-$\tau$ ACF, $\tau\in\{1,2,4,8\}$) \\
K (iterations) & 4 (MDD), 3 (FEP), 2 (PD) \\
Low-rank dimension $r$ & 1 (MDD/FEP), 2 (PD) \\
delta\_topk $m$ & 460 (top-half of $920$) \\
Amplitude bound $s_{\max}$ & 10 \\
$\lambda$ ($\ell_1$ on $\Delta h^K$) & 0.01 \\
Direction/amplitude MLP width & 64 \\
OP-CNN & $d_{\mathrm{model}}{=}64$; stem/block kernels $7/3$; dilations $(1,2,4,8)$ \\
Head (with FL-DPD) & MLP 42320$\to$256$\to$2 \\
Head (source) & linear 920$\to$2 \\
Batch size & 4 \\
Optimizer & AdamW, cosine-annealing LR schedule \\
Early stopping patience & 5 \\
\bottomrule
\end{tabularx}
\end{table}

The full diagnostic path is compact: the source branch has $168$k trainable parameters (OP-CNN $104$k; deformation solver and direction net $60$k; linear readout $1.8$k), the sensor expert adds its logistic-regression weights, and the total \dmdmodel{} model has $\approx169$k parameters (MDD $169{,}628$; FEP $168{,}623$; PD $169{,}592$). The $g_{\mathrm{norm}}$ development readout is excluded from the reported results.

\section{Ablations}\label{app:dev-ablation}

\begin{table}[t]
\caption{Solver depth and rank scan on the real-lag scale-axis (per-$\tau$ ACF) under the strict three-seed protocol (mean$\pm$std over seeds), including the one-shot ($K{=}1$) and no-deformation ($K{=}0$) controls. Candidates are the $K{\ge}2$ rows; the bold row is the inner-validation argmax over its disease's candidates. The control rows ($K{=}1$ and $K{=}0$) fall outside the candidate set. $^\ast$The no-deformation ($K{=}0$) row bypasses the solver entirely and routes the anchor state $h^0$ directly to the linear head, so it isolates the contribution of the iterative deformation beyond a zero-input control. The holdout columns are a post-selection check.}
\label{tab:ablation}
\centering
\small
\begin{tabular}{llccc}
\toprule
Disease & Solver & Inner-val AUC & Holdout AUC & Holdout BAcc \\
\midrule
MDD & $K{=}4, r{=}1$ & \textbf{0.801 $\pm$ 0.091} & 0.622 $\pm$ 0.134 & 0.622 $\pm$ 0.110 \\
MDD & $K{=}3, r{=}1$ & 0.794 $\pm$ 0.060 & 0.544 $\pm$ 0.096 & 0.550 $\pm$ 0.049 \\
MDD & $K{=}4, r{=}2$ & 0.785 $\pm$ 0.126 & 0.622 $\pm$ 0.063 & 0.567 $\pm$ 0.121 \\
MDD & $K{=}1, r{=}1$ & 0.785 $\pm$ 0.252 & 0.567 $\pm$ 0.119 & 0.506 $\pm$ 0.021 \\
MDD & $K{=}0$ (no deformation)$^\ast$ & 0.780 $\pm$ 0.085 & 0.600 $\pm$ 0.054 & 0.594 $\pm$ 0.031 \\
\midrule
FEP & $K{=}4, r{=}1$ & 0.719 $\pm$ 0.086 & 0.628 $\pm$ 0.031 & 0.567 $\pm$ 0.027 \\
FEP & $K{=}3, r{=}1$ & \textbf{0.826 $\pm$ 0.110} & 0.794 $\pm$ 0.139 & 0.706 $\pm$ 0.131 \\
FEP & $K{=}1, r{=}1$ & 0.796 $\pm$ 0.071 & 0.667 $\pm$ 0.142 & 0.611 $\pm$ 0.172 \\
FEP & $K{=}0$ (no deformation)$^\ast$ & 0.698 $\pm$ 0.108 & 0.678 $\pm$ 0.110 & 0.578 $\pm$ 0.075 \\
\midrule
PD & $K{=}4, r{=}1$ & 0.673 $\pm$ 0.051 & 0.552 $\pm$ 0.090 & 0.550 $\pm$ 0.061 \\
PD & $K{=}2, r{=}2$ & \textbf{0.680 $\pm$ 0.051} & 0.638 $\pm$ 0.020 & 0.608 $\pm$ 0.077 \\
PD & $K{=}1, r{=}1$ & 0.664 $\pm$ 0.050 & 0.500 $\pm$ 0.108 & 0.508 $\pm$ 0.031 \\
PD & $K{=}0$ (no deformation)$^\ast$ & 0.651 $\pm$ 0.034 & 0.623 $\pm$ 0.049 & 0.550 $\pm$ 0.041 \\
\bottomrule
\end{tabular}
\end{table}

With the support fixed at $m{=}460$, the solver depth and rank are selected per disease by inner validation (Table~\ref{tab:ablation}): MDD $K{=}4,r{=}1$, FEP $K{=}3,r{=}1$, PD $K{=}2,r{=}2$. The cohorts differ in subject count, montage, and sampling rate, so we set the depth and rank per dataset over a shared ROI--band--lag coordinate system. The scan is deliberately small---the shared low-rank subspace is a capacity prior for small cohorts ($r\le2$) and the deformation is a short refinement ($K\le4$)---and both controls stay below the selected depth on every task, so the iterative deformation carries the discriminative signal beyond a one-shot low-rank projection or the undeformed anchor state. The scanned candidates differ by at most about one seed standard deviation on the selection metric, and the inner-validation--holdout gap (e.g., MDD $0.801\to0.622$) is the optimism of selecting among a few configurations on a small inner pool; the reported value is the single holdout evaluation per seed.

\begin{table}[t]
\caption{Healthy-dynamics pretraining ablation under the strict three-seed protocol (mean$\pm$std over seeds). ``Full'' is the reported configuration with the disease's selected solver depth and rank (Table~\ref{tab:ablation}); ``scratch'' trains the same configuration from an empty pretrained checkpoint, so the healthy-dynamics prior is removed. The $K{=}1$ and $K{=}0$ deformation controls are in Table~\ref{tab:ablation}.}
\label{tab:components}
\centering\small
\begin{tabular}{llccc}
\toprule
Disease & Configuration & Inner-val AUC & Holdout AUC & Holdout BAcc \\
\midrule
MDD & full (healthy pretrained) & 0.801 $\pm$ 0.091 & 0.622 $\pm$ 0.134 & 0.622 $\pm$ 0.110 \\
MDD & scratch (no pretraining) & 0.756 $\pm$ 0.165 & 0.600 $\pm$ 0.027 & 0.578 $\pm$ 0.100 \\
\midrule
FEP & full (healthy pretrained) & 0.826 $\pm$ 0.110 & 0.794 $\pm$ 0.139 & 0.706 $\pm$ 0.131 \\
FEP & scratch (no pretraining) & 0.758 $\pm$ 0.052 & 0.778 $\pm$ 0.129 & 0.667 $\pm$ 0.095 \\
\midrule
PD & full (healthy pretrained) & 0.680 $\pm$ 0.051 & 0.638 $\pm$ 0.020 & 0.608 $\pm$ 0.077 \\
PD & scratch (no pretraining) & 0.634 $\pm$ 0.075 & 0.415 $\pm$ 0.076 & 0.508 $\pm$ 0.072 \\
\bottomrule
\end{tabular}
\end{table}

Clearing the pretrained checkpoint degrades the source branch on all three tasks, most sharply on PD: holdout AUC falls from $0.638$ to $0.415$ (PD), $0.794$ to $0.778$ (FEP), and $0.622$ to $0.600$ (MDD), with inner-validation AUC dropping on every task, so the healthy-dynamics prior stabilizes the learned deformation, doing more than initialization. The deformation's own contribution is the no-deformation ($K{=}0$) control in Table~\ref{tab:ablation}.

\subsection{Fusion variants}\label{app:fusion}

We also evaluated two adaptive alternatives to the fixed average: a global $\alpha$ scanned on the inner validation fold (F1), and logit-space fusion $z_{\mathrm{final}}=\alpha z_{\mathrm{sensor}}+(1-\alpha)z_{\mathrm{source}}$ with $\alpha$ chosen on the inner validation fold (F2). Neither reliably improved over the fixed average, and the selected $\alpha$ often collapsed to $0$ because the inner pool is too small to estimate a stable weight. We therefore keep the fixed $\alpha{=}0.5$ average, which also preserves the structural independence of the source state.

\section{Paired bootstrap of the fusion gain}

Table~\ref{tab:paired-dauc} reports the paired AUC difference between the fixed fusion and each single branch: fold-level values (descriptive) and subject-level values from a paired bootstrap over the pooled holdout subjects (20000 resamples, 95\% percentile CI, same resampling index for both). The PD fusion-minus-source gain has a 95\% CI that excludes zero.

\begin{table}[h]
\caption{Paired $\Delta$AUC of the $\alpha{=}0.5$ fusion against each single branch. Fold-level rows report mean$\pm$std over the three holdout folds (folds with positive $\Delta$AUC in parentheses). Subject-level rows report a paired bootstrap over the pooled holdout subjects ($n{=}33/48/90$; 20000 resamples; 95\% percentile CI). Bold marks the comparison whose 95\% CI excludes zero.}
\label{tab:paired-dauc}
\centering\small
\begin{tabular}{llcc}
\toprule
Disease & Level & $\Delta$AUC Fusion$-$Source & $\Delta$AUC Fusion$-$Sensor \\
\midrule
MDD & fold ($n{=}3$) & $+0.078 \pm 0.096$ (2/3) & $+0.078 \pm 0.107$ (2/3) \\
MDD & subject ($n{=}33$) & $+0.067$ [$-0.062$, $+0.198$] & $+0.052$ [$-0.070$, $+0.173$] \\
\midrule
FEP & fold ($n{=}3$) & $-0.006 \pm 0.139$ (1/3) & $+0.089 \pm 0.111$ (2/3) \\
FEP & subject ($n{=}48$) & $+0.030$ [$-0.104$, $+0.172$] & $+0.065$ [$-0.030$, $+0.164$] \\
\midrule
PD & fold ($n{=}3$) & $+0.142 \pm 0.026$ (3/3) & $+0.008 \pm 0.020$ (2/3) \\
PD & subject ($n{=}90$) & \textbf{$+0.176$ [$+0.048$, $+0.308$]} & $-0.006$ [$-0.049$, $+0.035$] \\
\bottomrule
\end{tabular}
\end{table}

\section{Fixed five-fold reference}

For ablation control we report fixed subject-level five-fold results in Table~\ref{tab:5fold5}. These values are selection-biased and the multi-seed holdout protocol replaces them as the trustworthy generalization estimate. Within this protocol, the best source-side configuration is the per-subject logit average of the two independently trained readouts, the direct-deformation and communication-pathway heads (MDD $0.763/0.712$; FEP $0.683/0.693$).

\begin{table}[h]
\caption{Fixed subject-level five-fold reference, mean$\pm$std. Not the primary estimate. Ensemble rows ($\dagger$) are per-subject logit averages of the two readouts. The strict-holdout results in the main text use the direct deformation readout. This 5-fold table additionally includes the communication-pathway readout as a development comparison. The no-deformation ($K{=}0$) control is reported in Table~\ref{tab:ablation}.}
\label{tab:5fold5}
\centering
\resizebox{\textwidth}{!}{%
\begin{tabular}{llccccc}
\toprule
Disease & Readout & Acc & BAcc & AUC & F1 \\
\midrule
MDD & with FL-DPD & 0.696 $\pm$ 0.111 & 0.705 $\pm$ 0.092 & 0.686 $\pm$ 0.158 & 0.676 $\pm$ 0.135 \\
MDD & Ours (source) & 0.642 $\pm$ 0.105 & 0.657 $\pm$ 0.109 & 0.698 $\pm$ 0.092 & 0.623 $\pm$ 0.116 \\
MDD & Ensemble$^\dagger$ & --- & 0.763$^\dagger$ & 0.712$^\dagger$ & --- \\
FEP & with FL-DPD & 0.719 $\pm$ 0.073 & 0.660 $\pm$ 0.097 & 0.669 $\pm$ 0.099 & 0.674 $\pm$ 0.111 \\
FEP & Ours (source) & 0.682 $\pm$ 0.084 & 0.634 $\pm$ 0.094 & 0.683 $\pm$ 0.088 & 0.649 $\pm$ 0.111 \\
FEP & Ensemble$^\dagger$ & --- & 0.683$^\dagger$ & 0.693$^\dagger$ & --- \\
PD & with FL-DPD & 0.671 $\pm$ 0.045 & 0.620 $\pm$ 0.083 & 0.598 $\pm$ 0.160 & 0.631 $\pm$ 0.059 \\
PD & Ours (source) & 0.624 $\pm$ 0.035 & 0.581 $\pm$ 0.043 & 0.613 $\pm$ 0.084 & 0.618 $\pm$ 0.034 \\
\bottomrule
\end{tabular}}
\end{table}

\section{Supplementary source-side analyses}

\subsection{FL-DPD communication-pathway ablation}

To test whether the frequency--lag directed pathway contributes to source-side classification, we zero the pathway-fusion weights at each 5-fold best checkpoint and rerun inference on the same validation subjects. This drops subject-level accuracy from $0.611\pm0.019$ to $0.491\pm0.046$ and AUC from $0.666\pm0.016$ to $0.500$, so the pathway carries most of the discriminative signal within this communication-pathway readout. The reported strict-holdout model reads the deformation $\Delta h^K$ directly (Section~\ref{sec:deform}), so this is a development comparison, separate from the reported decision path; it supports a predictive role for the pathway while leaving direction and lag semantics untested.

\subsection{Component-level diagnostics of the source model}

We examined which internal FL-DPD components train and carry information. In the earliest healthy-pretrained variants, the sparse gate (Hard-Concrete log-$\alpha$) stayed at its initialization (mean open rate $0.043$ over all $1{,}936$ gates), the cross-frequency term stayed near zero, and the scoring matrix $S$ was nearly uniform across bands, with no single edge significant under subject-level permutation (0/462). A label-shuffle permutation test (5 permutations) confirmed the classification signal is real: observed AUC $0.69$--$0.74$ versus null $0.458\pm0.055$. The source-side signal is therefore diffuse across edges and strongest in spectral power.

\subsection{Leakage correction for cross-ROI predictive increments}

A negative control asks whether cross-ROI delayed prediction reflects directed communication or zero-lag common-driver leakage. Raw cross-ROI incremental predictive ability ($\Delta R^2$, self AR(4) vs.\ self+cross AR(4)) was small ($+0.0116$ to $+0.0144$ across diseases) and vanished after symmetric orthogonalization ($-0.0003$), with the top-1 leakage component explaining $46$--$52\%$ of variance and the top-3 explaining about $100\%$ of the increment. Time reversal left this unchanged, and single-lag Granger increments and MVAR+PDC asymmetries also vanished. We therefore frame source-side communication as \emph{model-implied predictive communication}, distinct from biological directionality. The FL-DPD ablation establishes a predictive contribution, leaving causal pathway identity open.

\subsection{Cross-task state dependence of MDD deviations}

The cross-task experiment from resting to the dot-probe event-related-potential (ERP) task (same 53 MODMA subjects, resting-trained model) found no false-discovery-rate (FDR)-significant frequency, lag, or direction edges in either state (resting 6/462 nominal frequency edges, ERP 7/462; best ERP lag $p{=}0.409$). The resting network centered on DMN nodes (sgACC--PCC--dmPFC--amygdala), the ERP state on a cognitive-control pattern (DLPFC--amygdala--hippocampus--thalamus). The pathway-by-state interaction failed to reach significance ($t{=}-0.985$, $p{=}0.329$, $d{=}-0.269$, $n{=}53$), and masking predefined DMN or fronto-limbic edge sets changed AUC by less than $0.01$. These results support a preliminary state dependence, falling short of a verified dissociation, and constrain the single-state coordinate findings.

\subsection{External cross-cohort transfer}

The source branch is defined in the shared 46-ROI space and therefore uses the same source coordinate system across montages. The sensor expert and the fusion are tied to the recording montage and cannot be applied when the target cohort uses a different cap, so only the source branch is transferred. Table~\ref{tab:external} reports the three-seed transfer for MDD (MODMA $\to$ ds003478~\citep{ds003478openneuro}) and FEP (ds003944 $\to$ ds003947~\citep{ds003947openneuro}).

\begin{table}[h]
\caption{Cross-dataset transfer of the source branch (mean$\pm$std over the three seeds, each checkpoint evaluated once on the target cohort). The sensor expert and the fusion are tied to the recording montage and are excluded from transfer.}
\label{tab:external}
\centering\small
\begin{tabular}{llcc}
\toprule
Transfer & Target cohort ($n$) & AUC & BAcc \\
\midrule
MDD $\to$ ds003478 & 11 MDD / 75 HC & 0.724 $\pm$ 0.042 & 0.548 $\pm$ 0.073 \\
FEP $\to$ ds003947 & 31 psychosis / 30 control & 0.542 $\pm$ 0.032 & 0.543 $\pm$ 0.020 \\
\bottomrule
\end{tabular}
\end{table}

For MDD the ranking transfers (AUC $0.724\pm0.042$; group-mean gap $+0.093\pm0.029$, positive in all three seeds), though the fixed $0.5$ threshold mislabels many subjects (BAcc $0.548\pm0.073$), a calibration shift that leaves the signal intact. For FEP the group means are ordered correctly (gap $+0.045\pm0.024$) but the transfer AUC is near chance ($0.542\pm0.032$), so the ds003944 discrimination is cohort-specific and we make no cross-cohort generalization claim for FEP. Both cohorts are small and the folded metrics are class-balance sensitive, so only the ranking AUC is used as transfer evidence.

\subsection{Additional ablations}

Temporal position embeddings and attention-versus-mean pooling showed no reliable contribution: training only the position embedding and head changed MDD accuracy by $+0.056\pm0.046$ and AUC by $-0.007\pm0.013$ (3/5 folds), and attention pooling was indistinguishable from mean pooling (AUC $+0.013\pm0.027$). In the FEP fine-tuning grid, the observed degradation traced to the focal-loss exponent ($\gamma{=}2\to0$), while the communication loss weight stayed unchanged (fold-1 AUC $0.833\to0.500$). Discriminative power is thus concentrated in the spectral and deformation paths, with patch order and pooling contributing little.

\subsection{Lag-axis validity: synthetic recovery and lag perturbation}\label{app:lag}

The lag axis is the real per-$\tau$ autocorrelation of the op-CNN latent state ($\tau\in\{1,2,4,8\}$), so its validity is established at the method level, independent of matching a lag to a literature value. Two experiments test the axis in opposite directions: a planted-lag recovery (positive control, whether the representation can encode a known lag) and a source-level lag perturbation (necessity, whether the decision uses it). The perturbations act on the solver's input state $h_{\mathrm{health}}$ of the chosen-fold checkpoint on the fold-validation subjects, and we report the clip-level AUC change relative to the identity perturbation.

\paragraph{Synthetic recovery.}
A narrowband oscillation at frequency $f_{\mathrm{hp}}/\ell^\star$ is planted in one target source channel so that its autocorrelation peaks at $\ell^\star\in\{1,2,4,8\}$, and a logistic probe on the four-lag ACF predicts $\ell^\star$ with a half/half train/test split. Both the op-CNN latent ACF and the raw ACF recover the planted lag at accuracy $1.00$ (chance $0.25$; Table~\ref{tab:synth-lag}), confirming that the lag coordinates encode a known lag.

\begin{table}[h]
\caption{Synthetic lag recovery (positive control): a logistic probe on the four-lag ACF recovers the planted lag $\ell^\star\in\{1,2,4,8\}$ from either the op-CNN latent or the raw ACF (train/test split in half).}
\label{tab:synth-lag}
\centering\small
\begin{tabular}{lcc}
\toprule
Probe input & Accuracy & Chance \\
\midrule
op-CNN latent ACF & $1.00$ & $0.25$ \\
Raw ACF & $1.00$ & $0.25$ \\
\bottomrule
\end{tabular}
\end{table}

\paragraph{Lag perturbation.}
Table~\ref{tab:lag-pert} reports the three-seed mean $\Delta$AUC at the solver input. Replacing the per-$\tau$ ACF by a time mean (the abstract-lag condition) gives the largest drop (MDD $-0.160$, FEP $-0.138$, PD $-0.086$), the same ordering as the real-lag versus abstract-lag comparison; permuting or rolling the $\tau$ anchoring gives smaller drops ($-0.03$ to $-0.06$). In the incremental test, lag adds beyond band on MDD ($-0.034$) and PD ($-0.047$), and fails to add on FEP ($+0.006$). The band and ROI perturbations are positive controls for the other axes.

\begin{table}[h]
\caption{Source-level lag perturbation at the solver input (three-seed mean $\Delta$AUC relative to the identity perturbation; chosen-fold checkpoint on fold-validation subjects). The incremental rows add lag on top of band or ROI, with the increment over the single-axis perturbation in parentheses.}
\label{tab:lag-pert}
\centering\small
\begin{tabular}{lccc}
\toprule
Perturbation & MDD & FEP & PD \\
\midrule
$\tau$ anchoring permuted & $-0.043$ & $-0.049$ & $-0.052$ \\
$\tau$ anchoring rolled & $-0.063$ & $-0.034$ & $-0.062$ \\
per-$\tau$ ACF $\to$ time mean & $-0.160$ & $-0.138$ & $-0.086$ \\
full time shuffle & $-0.074$ & $-0.101$ & $-0.163$ \\
\midrule
band permuted (control) & $-0.073$ & $-0.132$ & $-0.024$ \\
ROI permuted (control) & $-0.124$ & $-0.177$ & $-0.079$ \\
\midrule
lag $+$ band & $-0.106$ ($-0.034$) & $-0.126$ ($+0.006$) & $-0.072$ ($-0.047$) \\
lag $+$ ROI & $-0.157$ ($-0.034$) & $-0.165$ ($+0.011$) & $-0.078$ ($+0.001$) \\
\bottomrule
\end{tabular}
\end{table}

The ACF is even in $\tau$, so time reversal and time roll leave the four-lag features invariant ($\cos=1.00$, $\Delta$AUC $\approx0$). The lag axis therefore encodes the magnitude $|\tau|$, leaving direction unsigned. A directed-lag claim would require an asymmetric lag feature.

\subsection{Attribution coordinates}\label{app:attribution}

The consolidated highest-importance coordinates discussed in the main text are listed in Table~\ref{tab:attribution}.

\begin{table*}[t]
\caption{\textbf{Attribution readout of the healthy-anchored state $h_{\mathrm{health}}$.} ROI importance is the unsigned amplitude saliency $\mathrm{mean}_{\mathrm{subj}}|\partial\,\mathrm{logit}/\partial h_{\mathrm{health}}|$ (consensus over 3 splits $\times$ 3 folds), cross-seed stable (axis cosine $0.83$--$0.87$) and concordant with the occlusion test below (ROI Spearman $+0.31$ MDD, $+0.59$ PD). The spectro-temporal signature is the ROI-marginal signed saliency profile, carrying band and lag direction and selectivity. Each coordinate sits on a real axis: ROI and band inherited from the input, lag the real per-$\tau$ autocorrelation ($\tau\in\{1,2,4,8\}$). A band/lag $(+)$ sign means the patient-versus-healthy logit rises as that coordinate increases, marking a direction of decision sensitivity, distinct from a change in brain activity. Directional saliency findings.}
\label{tab:attribution}
\centering
\small
\setlength{\tabcolsep}{6pt}
\renewcommand{\arraystretch}{1.25}
\begin{tabularx}{\textwidth}{
    >{\raggedright\arraybackslash}p{0.07\textwidth}
    >{\raggedright\arraybackslash}p{0.40\textwidth}
    >{\raggedright\arraybackslash}X
}
\toprule
\textbf{Disease} &
\textbf{High-importance ROIs (amplitude)} &
\textbf{Spectro-temporal signature (signed)} \\
\midrule
\textbf{MDD} &
MTG\_R, DLPFC\_R, parahippocampal\_L, amygdala\_R, dACC\_R, dmPFC\_L, sgACC\_L, PCC\_L &
\textbf{Band:} $\delta$ $-1.40$ $\cdot$ $\theta$ $+0.73$ $\cdot$ $\alpha$ $+0.77$ $\cdot$ $\beta$ $-0.72$ $\cdot$ $\gamma$ $+0.33$

\textbf{Lag:} $\tau_1$ $+0.57$ $\cdot$ $\tau_2$ $-0.59$ $\cdot$ $\tau_4$ $+0.02$ $\cdot$ $\tau_8$ $-0.29$ \\
\midrule
\textbf{FEP} &
IFG\_R, visual\_association\_R, precuneus\_R, SPL\_L, insula\_R, IPL\_angular\_R, dmPFC\_R, sgACC\_L &
\textbf{Band:} $\delta$ $+0.26$ $\cdot$ $\theta$ $-0.24$ $\cdot$ $\alpha$ $-0.45$ $\cdot$ $\beta$ $+0.31$ $\cdot$ $\gamma$ $-0.20$

\textbf{Lag:} $\tau_1$ $-0.84$ $\cdot$ $\tau_2$ $-0.12$ $\cdot$ $\tau_4$ $-0.03$ $\cdot$ $\tau_8$ $+0.66$ \\
\midrule
\textbf{PD} &
premotor\_L, IPL\_angular\_R, IPL\_angular\_L, S1\_L, M1\_L, lateral\_OFC\_R, SMA\_R, precuneus\_L &
\textbf{Band:} $\delta$ $-0.13$ $\cdot$ $\theta$ $+0.01$ $\cdot$ $\alpha$ $+0.75$ $\cdot$ $\beta$ $-1.58$ $\cdot$ $\gamma$ $+0.01$

\textbf{Lag:} $\tau_1$ $-0.13$ $\cdot$ $\tau_2$ $-0.14$ $\cdot$ $\tau_4$ $-0.11$ $\cdot$ $\tau_8$ $-0.55$ \\
\bottomrule
\end{tabularx}
\end{table*}

\subsection{ROI occlusion on the decision}\label{app:occlusion}

Table~\ref{tab:occlusion} occludes the top-ranked ROIs and joint ROI sets: the set's full band~$\times$~lag block(s) in $h_{\mathrm{health}}$ are zeroed and the holdout re-scored. Single ROIs are ranked by occlusion magnitude over all 46 ROIs, independent of saliency; joint sets ($\dagger$) are occlusion-search results, reported as exploratory.

\begin{table}[h]
\caption{ROI occlusion on the healthy-anchored state. AUC drop (mean$\pm$std over the three seed checkpoints, chosen fold) when the occluded set's band~$\times$~lag block is zeroed and the holdout is re-scored. Single ROIs are ranked by occlusion magnitude over all 46 ROIs; joint sets ($\dagger$) are occlusion-search results and exploratory. The permutation $p$ over random same-size block sets (internal null, $n{=}300$) is nominal and uncorrected for the search over 46 ROIs, so the table is a descriptive ranking only.}
\label{tab:occlusion}
\centering
\small
\begin{tabular}{lll}
\toprule
Disease & Occluded set & $\Delta$AUC $\pm$ std ($p$) \\
\midrule
FEP & SPL\_L & $+0.056 \pm 0.042$ (0.003) \\
    & SPL\_R & $+0.050 \pm 0.049$ (0.004) \\
    & dmPFC\_L & $+0.050 \pm 0.014$ (0.005) \\
    & insula\_R & $+0.044 \pm 0.028$ (0.035) \\
    & M1\_R & $+0.039 \pm 0.016$ (0.017) \\
    & thalamus\_R & $+0.033 \pm 0.036$ (0.018) \\
    & visuo\_motor6$^\dagger$ & $+0.106$ (0.040) \\
\midrule
PD & premotor\_L & $+0.028 \pm 0.016$ (0.004) \\
   & M1\_L & $+0.023 \pm 0.024$ (0.010) \\
   & dACC\_L & $+0.017 \pm 0.010$ (0.028) \\
   & motor\_insula3$^\dagger$ & $+0.030$ (0.060) \\
\midrule
MDD & dACC\_R & $+0.044 \pm 0.016$ (0.045) \\
    & motor\_dacc6$^\dagger$ & $+0.067$ (0.020) \\
    & limbic6$^\dagger$ & $+0.056$ (0.060) \\
\bottomrule
\end{tabular}

\vspace{2pt}
\footnotesize $^\dagger$\,motor\_dacc6 $=$ \{IFG\_L, M1\_L, SMA\_L, amygdala\_R, dACC\_R, premotor\_L\}; limbic6 $=$ \{DLPFC\_L, amygdala\_R, dACC\_L, hippocampus\_L, parahippocampal\_L, thalamus\_L\}; visuo\_motor6 $=$ \{SPL\_L, SPL\_R, STG\_auditory\_R, premotor\_L, premotor\_R, sgACC\_R\}; motor\_insula3 $=$ \{insula\_L, precuneus\_R, premotor\_L\}.
\end{table}

\section{Supplementary figures}

These figures are descriptive (only the AUROC panels give unbiased performance estimates) and show \emph{where and in which direction} the source state differs.

\subsection{ROI $\times$ frequency $\times$ lag effect maps}

\begin{figure}[tb]
\centering
\begin{subfigure}{0.32\textwidth}
\includegraphics[width=\textwidth]{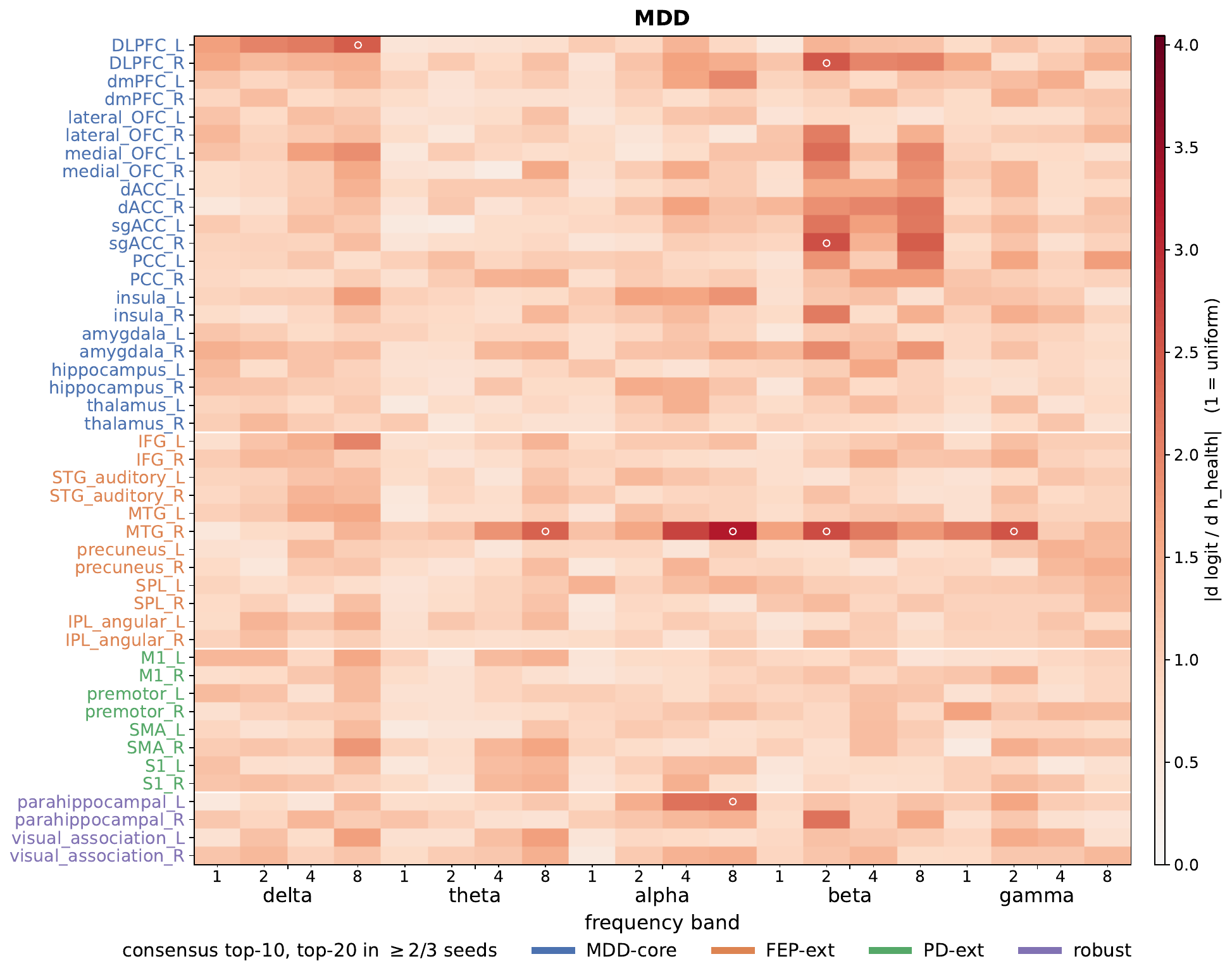}
\caption{MDD ($K{=}4,r{=}1$).}
\label{fig:attrib-mdd}
\end{subfigure}\hfill
\begin{subfigure}{0.32\textwidth}
\includegraphics[width=\textwidth]{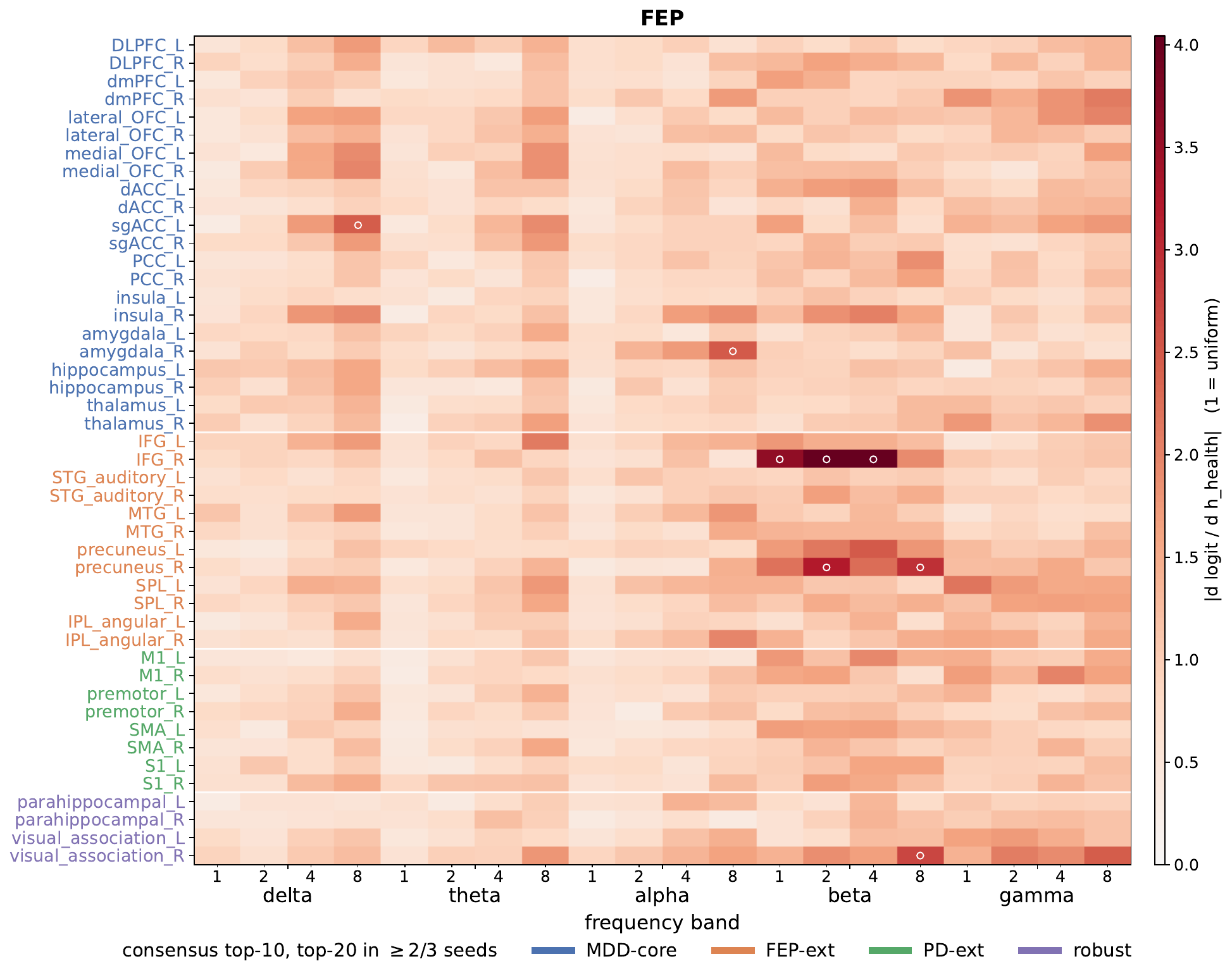}
\caption{FEP ($K{=}3,r{=}1$).}
\label{fig:attrib-fep}
\end{subfigure}\hfill
\begin{subfigure}{0.32\textwidth}
\includegraphics[width=\textwidth]{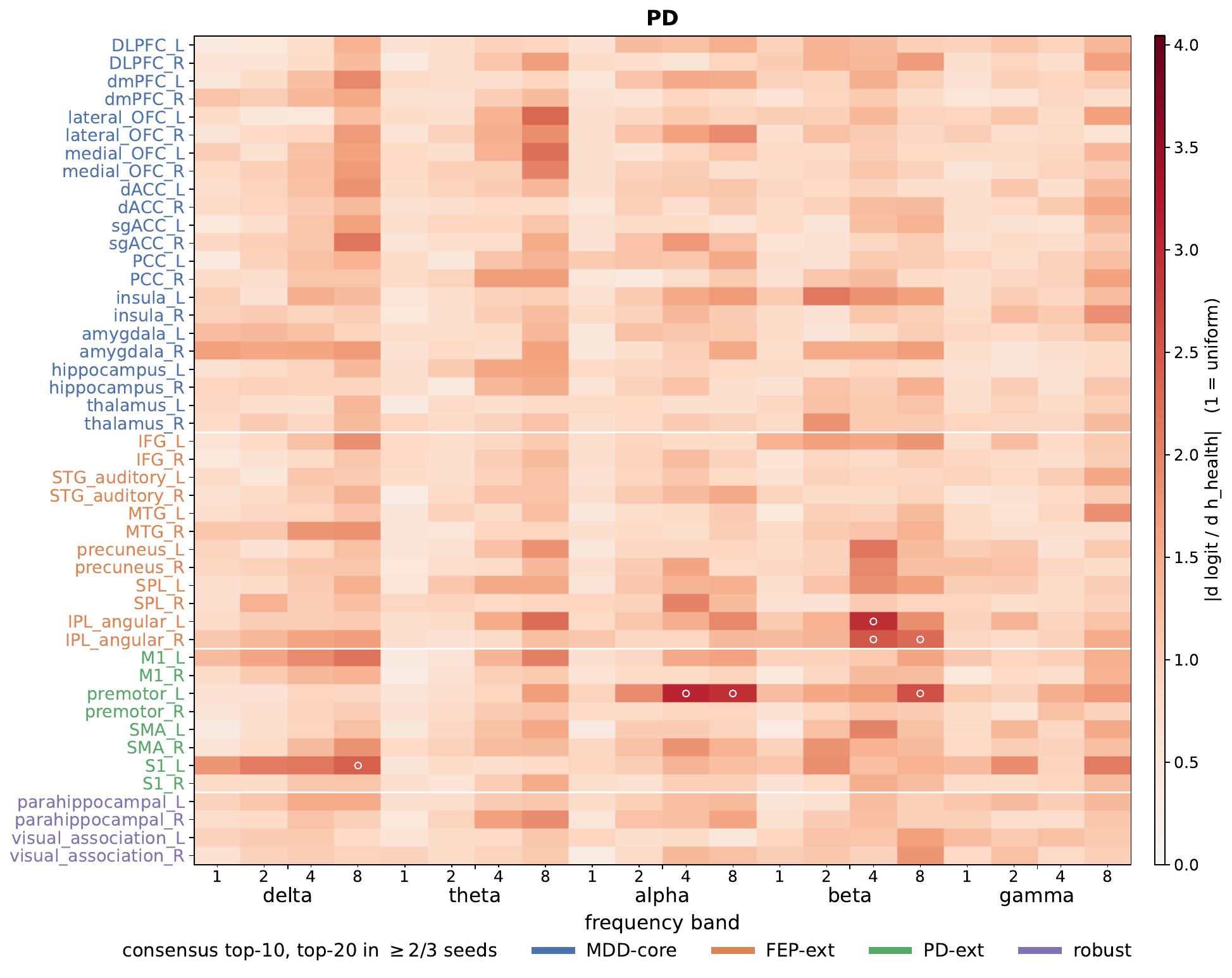}
\caption{PD ($K{=}2,r{=}2$).}
\label{fig:attrib-pd}
\end{subfigure}
\caption{\textbf{Decision-sensitivity attribution maps} of the healthy-anchored state (same $46$ ROI $\times$ $5$ band $\times$ $4$ lag grid as Figure~\ref{fig:effmap}). Color is the 3-seed consensus unsigned sensitivity $\mathrm{mean}_{\mathrm{subj}}|\partial\,\mathrm{logit}/\partial h_{\mathrm{health}}|$, normalized so $1.0$ is uniform (white $=0$, deeper red $=$ more sensitive); circles mark coordinates in the consensus top-10 and in $\ge2/3$ seeds' top-20. This map asks \emph{where the decision is sensitive}, Figure~\ref{fig:effmap} \emph{where the state differs}; the two are complementary. ROI-level attribution is cross-seed stable (axis cosine $0.83$--$0.87$), cell-level values only weakly stable, and sensitivity differs from decision-dependence (Table~\ref{tab:occlusion}).}
\label{fig:attrib}
\end{figure}

\begin{figure}[tb]
\centering
\begin{subfigure}{0.32\textwidth}
\includegraphics[width=\textwidth]{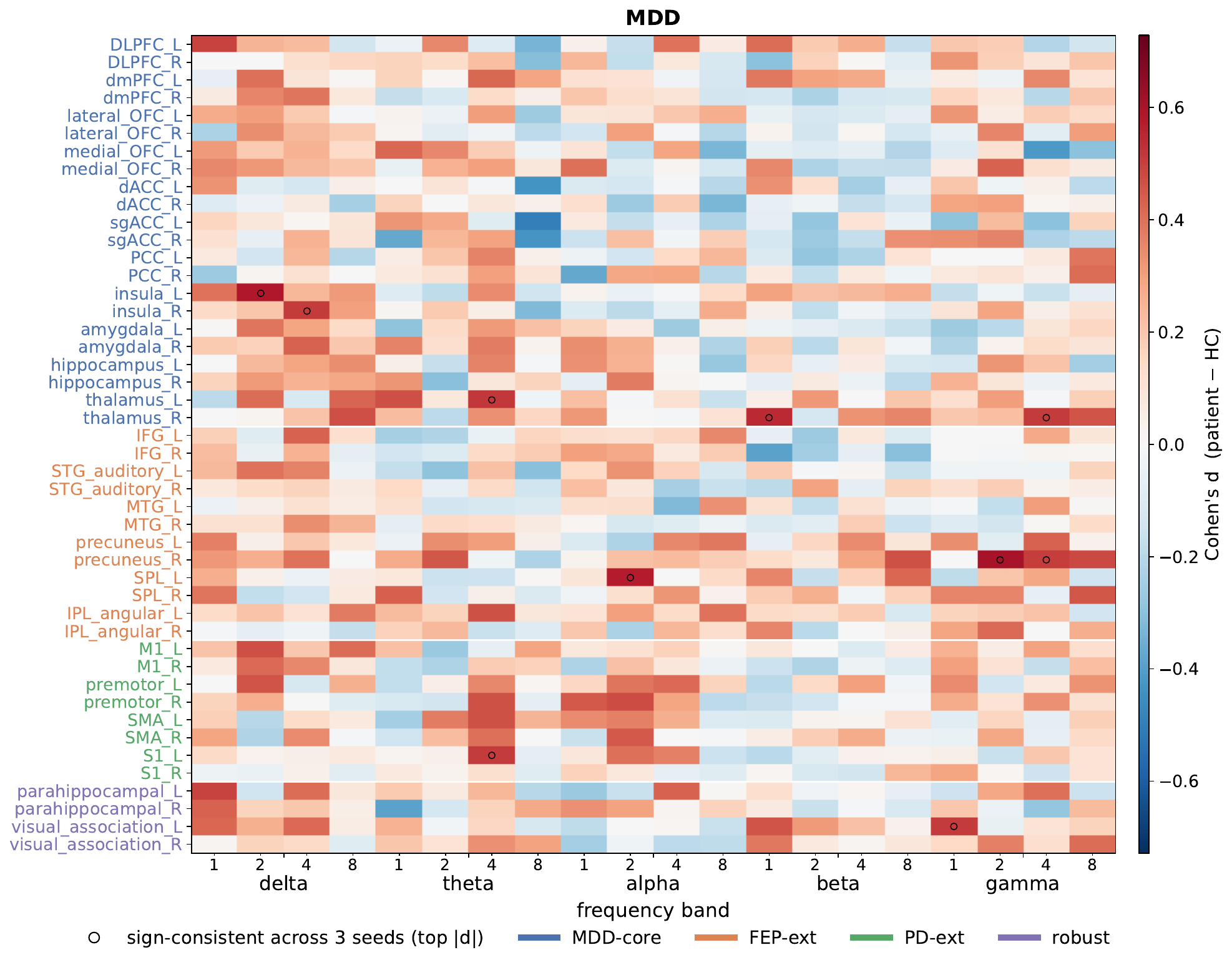}
\caption{MDD ($K{=}4,r{=}1$).}
\label{fig:effmap-mdd}
\end{subfigure}\hfill
\begin{subfigure}{0.32\textwidth}
\includegraphics[width=\textwidth]{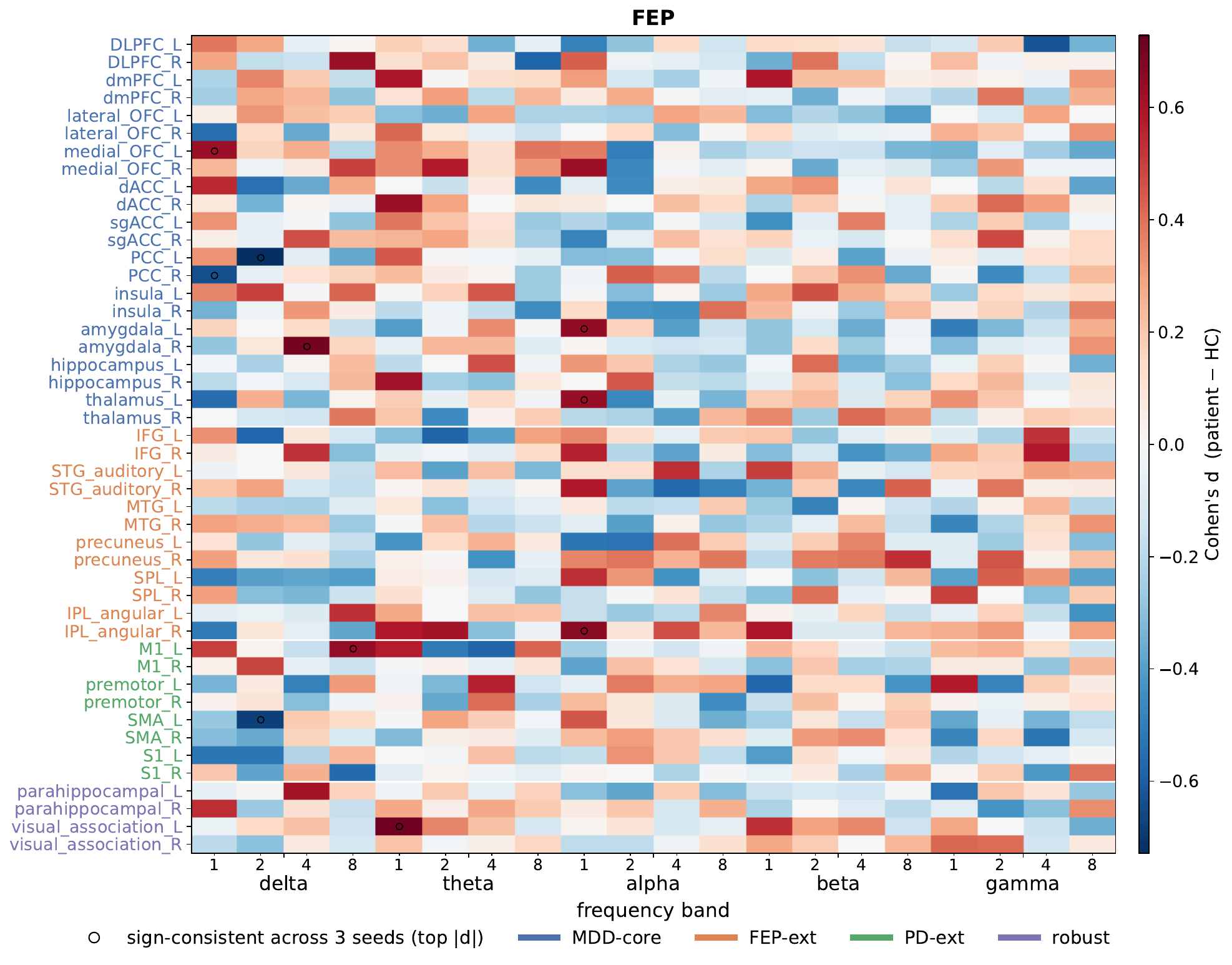}
\caption{FEP ($K{=}3,r{=}1$).}
\label{fig:effmap-fep}
\end{subfigure}\hfill
\begin{subfigure}{0.32\textwidth}
\includegraphics[width=\textwidth]{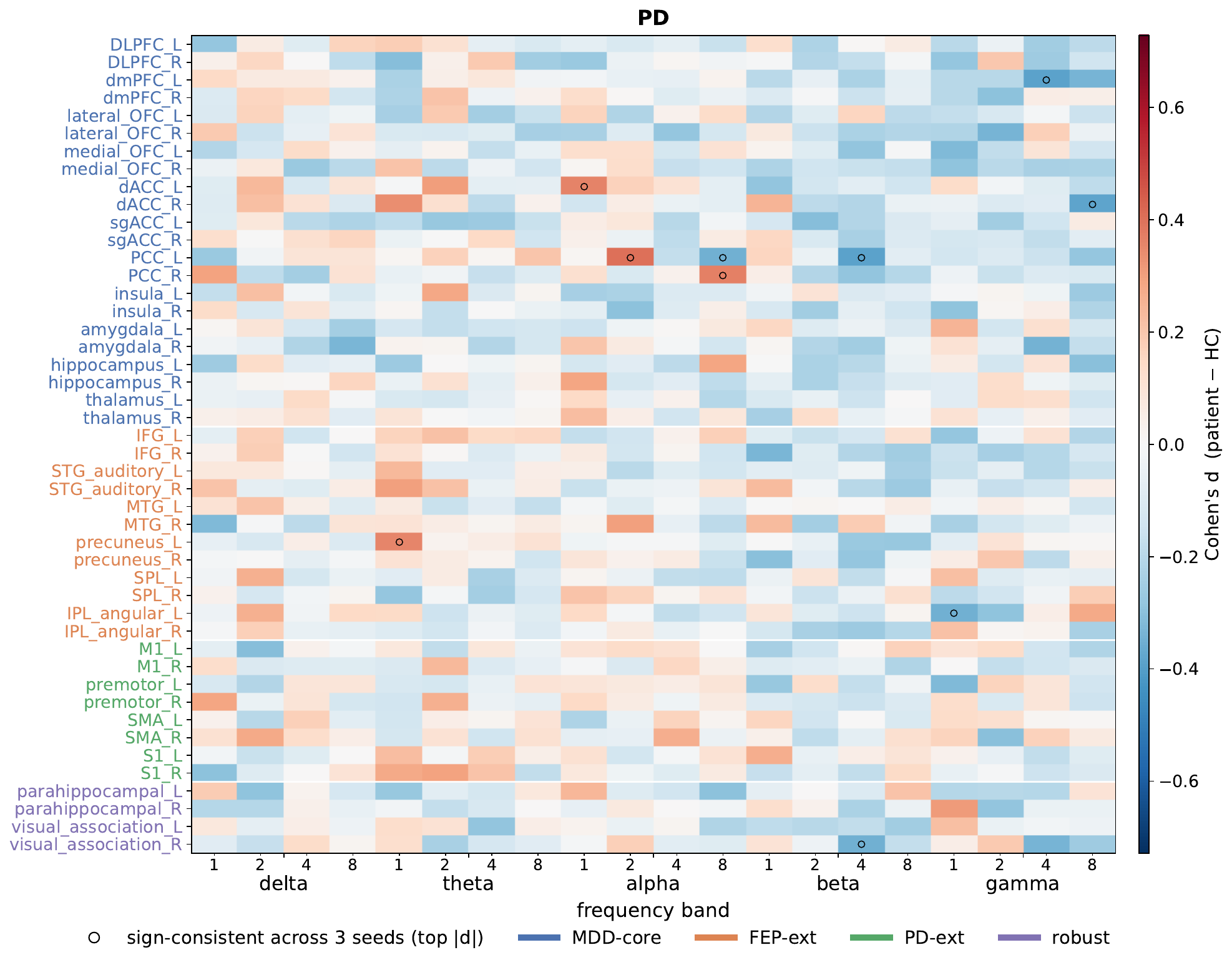}
\caption{PD ($K{=}2,r{=}2$).}
\label{fig:effmap-pd}
\end{subfigure}
\caption{\textbf{Source-state effect maps} of the final deformed state $h^{K}$ ($46$ ROI $\times$ $5$ bands $\times$ $4$ lags), same 3-seed full-cohort protocol as Figure~\ref{fig:attrib}. Rows are the $46$ ROIs grouped by atlas origin (MDD-core 22 / FEP-ext 12 / PD-ext 8 / robust 4; white lines separate groups); columns are the five bands, each split into lags $\{1,2,4,8\}$. Color is the mean Cohen's $d$ (patient $-$ HC) of the per-subject final state on a fixed $\pm1.0$ axis (red $=$ patients more positive, blue $=$ more negative). Circles mark coordinates whose $d$ sign is consistent across seeds and whose $|d|$ is in the top-$k$. Descriptive effect maps, distinct from significance maps.}
\label{fig:effmap}
\end{figure}

These maps read the source state along its real axes (ROI and band inherited from the input, lag from the real per-$\tau$ ACF), so each of the $46\times5\times4$ coordinates is directly readable. The decision-sensitivity attribution (Figure~\ref{fig:attrib}) is primary; the state-difference map (Figure~\ref{fig:effmap}) is a descriptive supplement.

Figure~\ref{fig:effmap} shows how the state itself differs between patients and HC (mean Cohen's $d$ of $h^{K}$ over the same grid), complementing the decision-sensitivity map. The two quantify different properties and can take different patterns; their low cell-level correlation (de-meaned Pearson $-0.02$ to $-0.11$) reflects this difference.

\subsection{Per-subject deformation trajectory and state space}

These panels show whether a subject's deviation accumulates through iterative solving and whether the groups occupy different state-space regions and in which direction the decision pushes each subject.

\begin{figure}[tb]
\centering
\begin{subfigure}{0.62\textwidth}
\includegraphics[width=\textwidth]{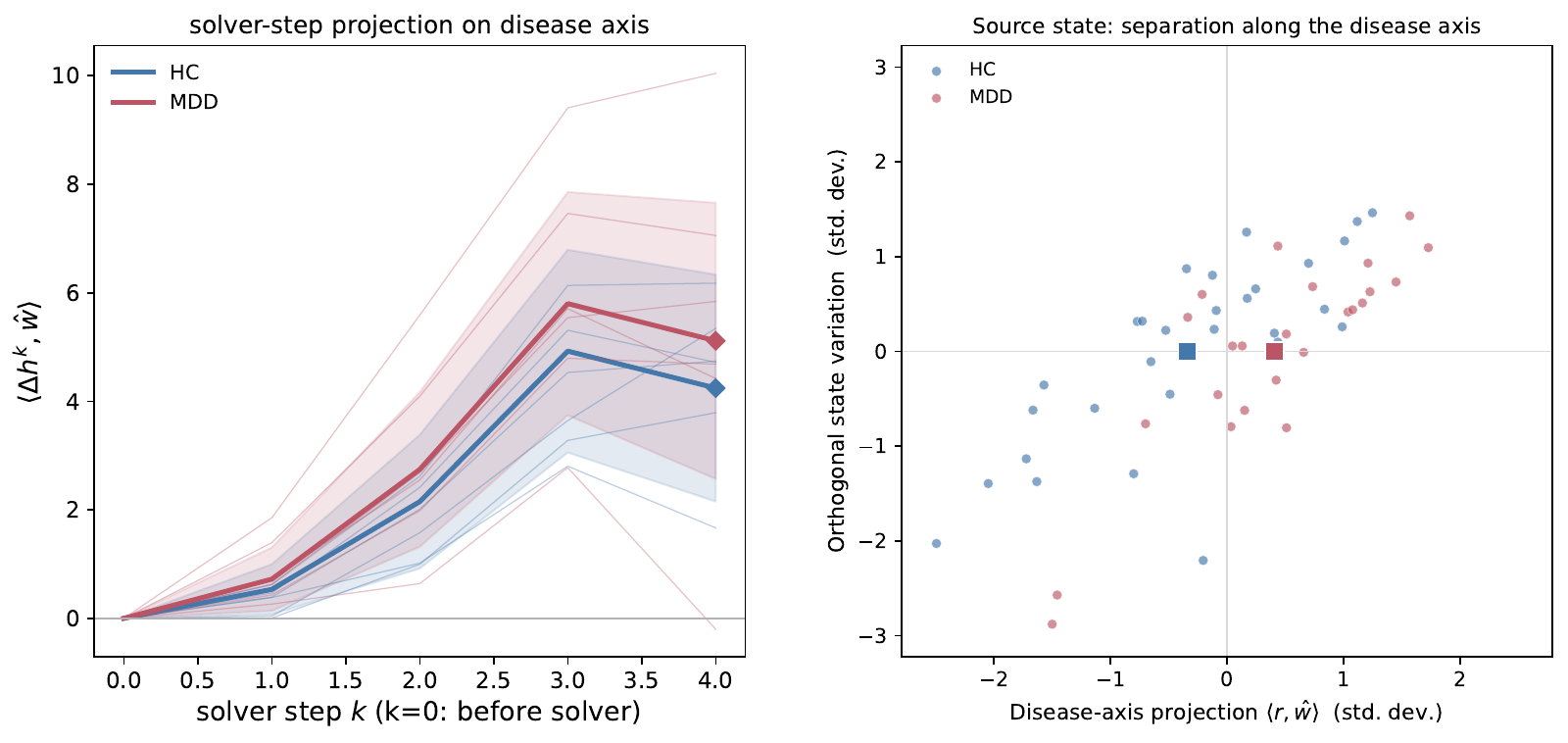}
\caption{MDD.}
\label{fig:proj-mdd}
\end{subfigure}

\begin{subfigure}{0.62\textwidth}
\includegraphics[width=\textwidth]{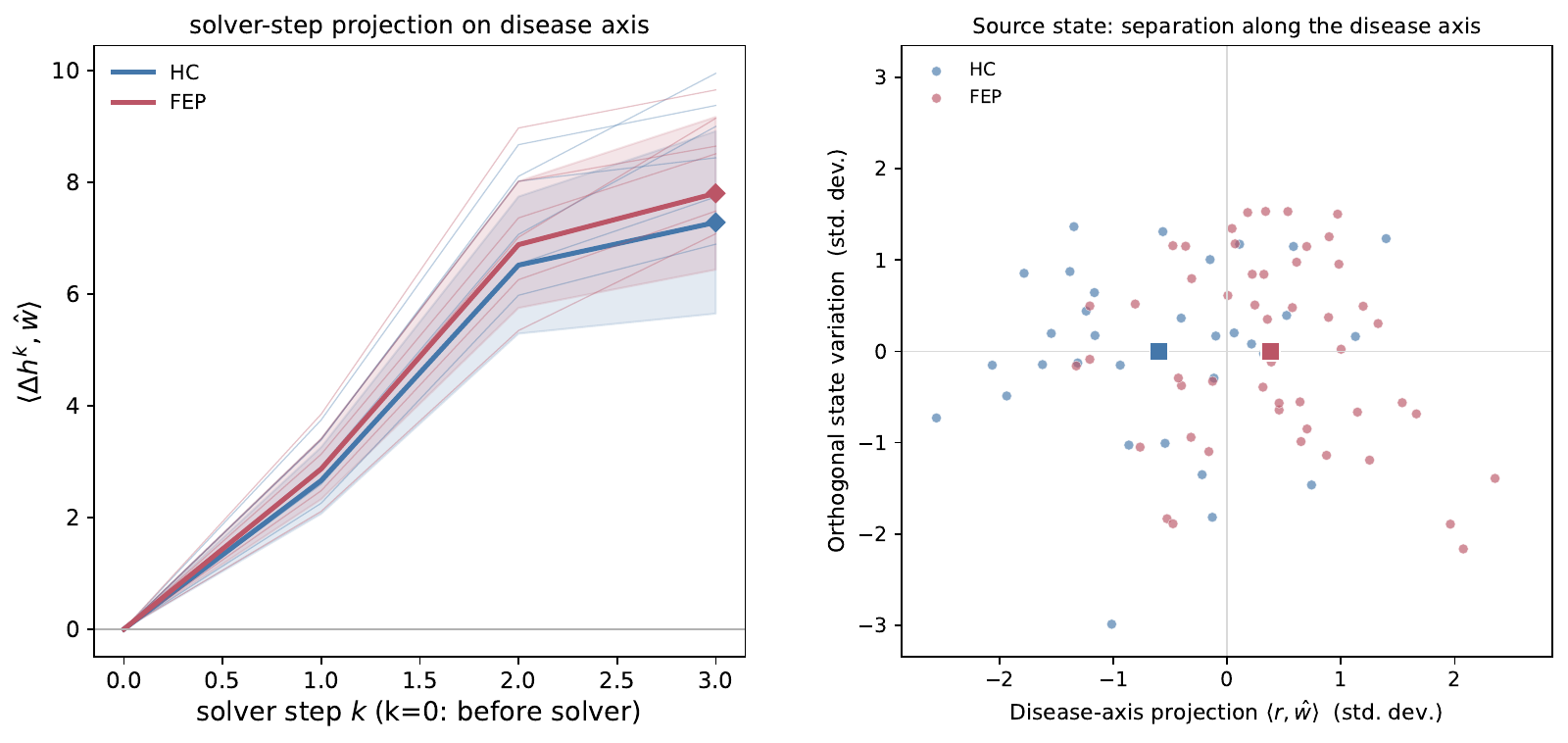}
\caption{FEP.}
\label{fig:proj-fep}
\end{subfigure}

\begin{subfigure}{0.62\textwidth}
\includegraphics[width=\textwidth]{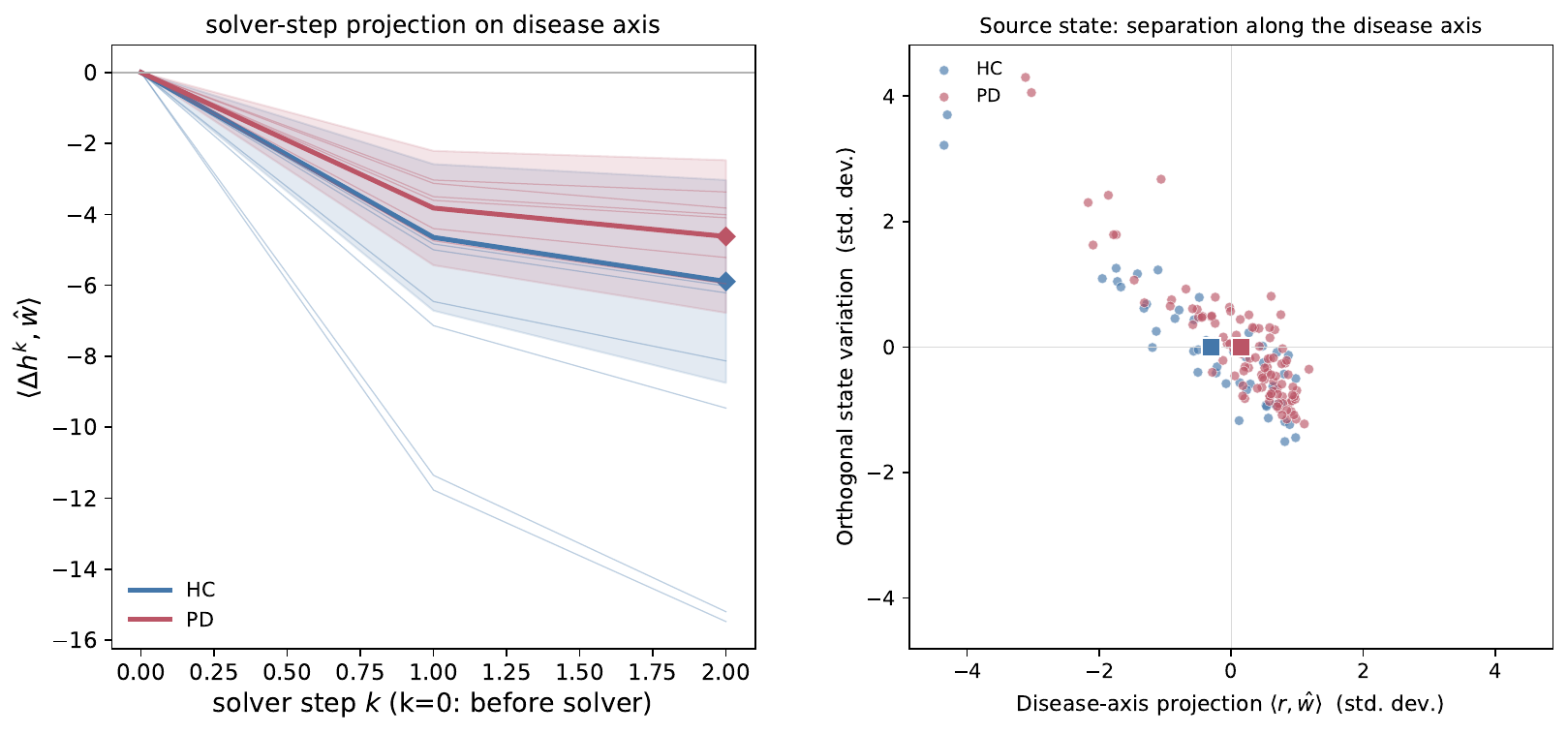}
\caption{PD.}
\label{fig:proj-pd}
\end{subfigure}
\caption{\textbf{Per-subject deformation trajectory and state geometry} (descriptive; one representative checkpoint per panel, cv-pool subjects seen in training). Left: the disease-axis projection $\langle\Delta h^k,\hat w\rangle$ as the solver iterates $k$ (thin lines, subjects; solid line, group mean $\pm$ std); monotone group separation indicates the deviation emerges during solving. Right: the two directions of largest cross-subject variance of $h_{\mathrm{health}}$ (920-dim) define a shared 2D plane with per-subject points (HC blue, patients red), group-mean squares, one HC$\to$patient offset arrow (direction real; length scaled for visibility, offset in $\sigma$ annotated), and per-subject quiver arrows of the decision gradient $\partial\,\mathrm{logit}/\partial h_{\mathrm{health}}$ (direction $=$ where the state must move to raise the patient probability; length $=$ sensitivity). Axes are normalized per principal direction to $\sigma$ units on a fixed $\pm3.5\sigma$ range shared across diseases.}
\label{fig:proj}
\end{figure}

\subsection{Dual-view AUROC}

These curves compare the ranking ability of the sensor, source, and fusion on pooled holdout subjects, separating ranking (AUC) from the operating point (BAcc/F1).

\begin{figure}[tb]
\centering
\begin{subfigure}{0.32\textwidth}
\includegraphics[width=\textwidth]{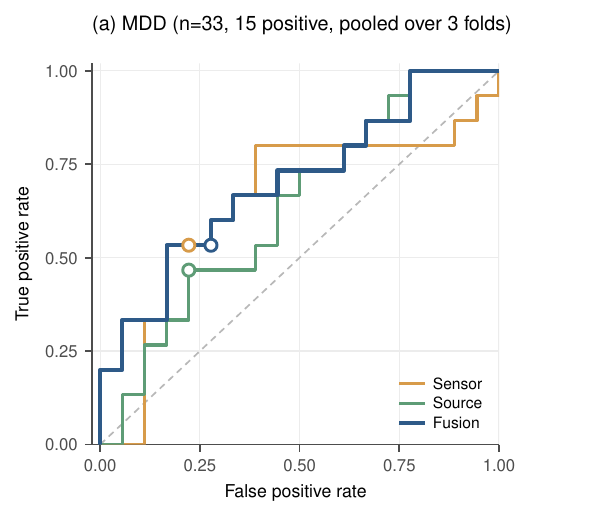}
\caption{MDD.}
\label{fig:auroc-mdd}
\end{subfigure}\hfill
\begin{subfigure}{0.32\textwidth}
\includegraphics[width=\textwidth]{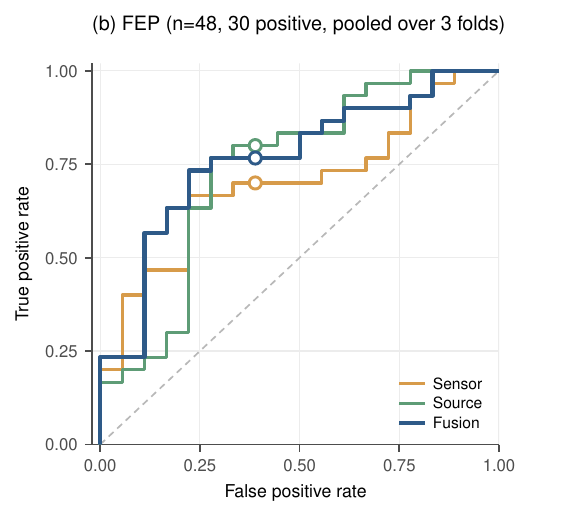}
\caption{FEP.}
\label{fig:auroc-fep}
\end{subfigure}\hfill
\begin{subfigure}{0.32\textwidth}
\includegraphics[width=\textwidth]{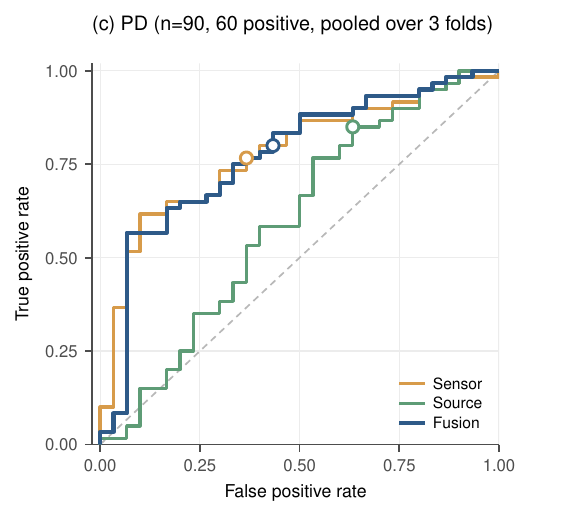}
\caption{PD.}
\label{fig:auroc-pd}
\end{subfigure}
\caption{\textbf{Receiver operating characteristic of the Sensor Expert, Source Expert, and \dmdmodelFull{}} on the pooled subjects of the three non-overlapping holdouts (each subject predicted once; $n{=}33$ MDD, $48$ FEP, $90$ PD). Fusion in dark blue, Sensor in amber, Source in sage green; open circles mark the fixed $0.5$-threshold operating point that determines BAcc/F1. AUC here is the pooled value over concatenated holdout subjects and differs from the per-seed mean$\pm$std in Tables~\ref{tab:dualview} and~\ref{tab:strict-baselines}. Fusion is competitive with and complementary to the best single branch: on MDD a positive trend over both branches, on FEP matching the strong source branch, on PD exceeding the source branch while matching the sensor. Fusion and sensor can share the $0.5$-threshold point, hence the same BAcc, while differing in AUC; the ROC separates ranking quality from the operating point. Curves are piecewise linear because the empirical ROC has finitely many thresholds.}
\label{fig:auroc}
\end{figure}

\subsection{Confidence scatter}

These plots show whether the branches make the same or different errors on individual subjects; different mistakes (points off the $y{=}x$ line in the disagreement quadrants) are the precondition for fusion to help.

\begin{figure}[tb]
\centering
\begin{subfigure}{0.30\textwidth}
\includegraphics[width=\textwidth]{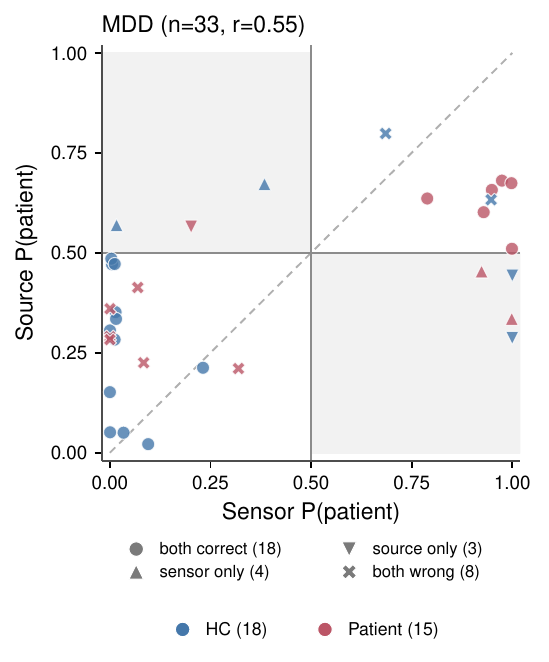}
\caption{MDD.}
\label{fig:scatter-mdd}
\end{subfigure}\hfill
\begin{subfigure}{0.30\textwidth}
\includegraphics[width=\textwidth]{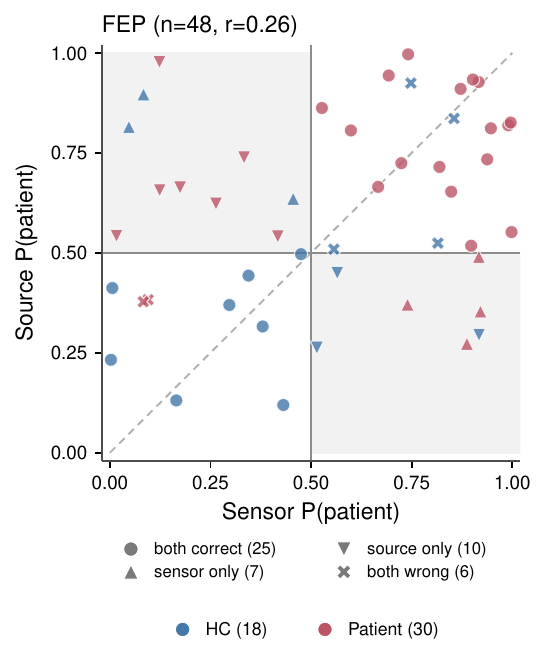}
\caption{FEP.}
\label{fig:scatter-fep}
\end{subfigure}\hfill
\begin{subfigure}{0.30\textwidth}
\includegraphics[width=\textwidth]{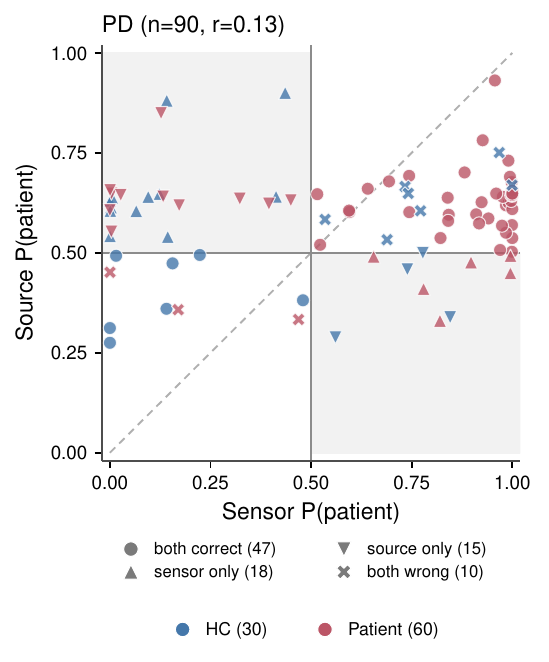}
\caption{PD.}
\label{fig:scatter-pd}
\end{subfigure}
\caption{\textbf{Sensor versus Source predicted probabilities} ($P(\text{patient})$; HC blue, patients red). Shapes encode which branch is correct: $\circ$ both, $\triangle$ sensor only, $\triangledown$ source only, $\times$ both wrong (counts in legend). The $0.5$ threshold lines define four quadrants; the two disagreement quadrants are shaded and the dashed $y{=}x$ line marks agreement. Pooled Pearson correlations (on the panels): MDD $r{=}0.55$, FEP $0.26$, PD $0.13$. Points in the disagreement triangles are subjects one branch gets wrong and the other recovers; sensor-only/source-only counts are $4/3$ (MDD), $7/10$ (FEP), $18/15$ (PD). On PD both branches err on many subjects and the source branch is weaker, so the fusion stays sensor-dominated.}
\label{fig:scatter}
\end{figure}

\subsection{Disease pathway glass brains}

These glass brains show which source-space ROIs the decision is most sensitive to in each disease and their anatomical arrangement.

\begin{figure*}[tb]
\centering
\begin{subfigure}{0.62\textwidth}
\includegraphics[width=\textwidth]{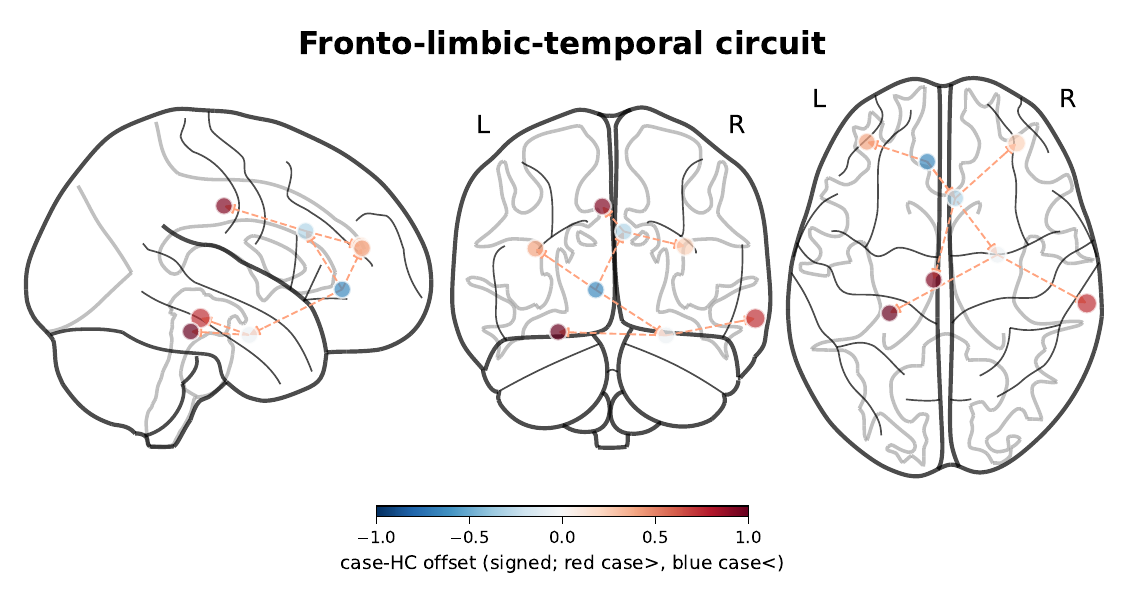}
\caption{MDD (fronto-limbic-temporal).}
\label{fig:glassbrain-mdd}
\end{subfigure}

\begin{subfigure}{0.62\textwidth}
\includegraphics[width=\textwidth]{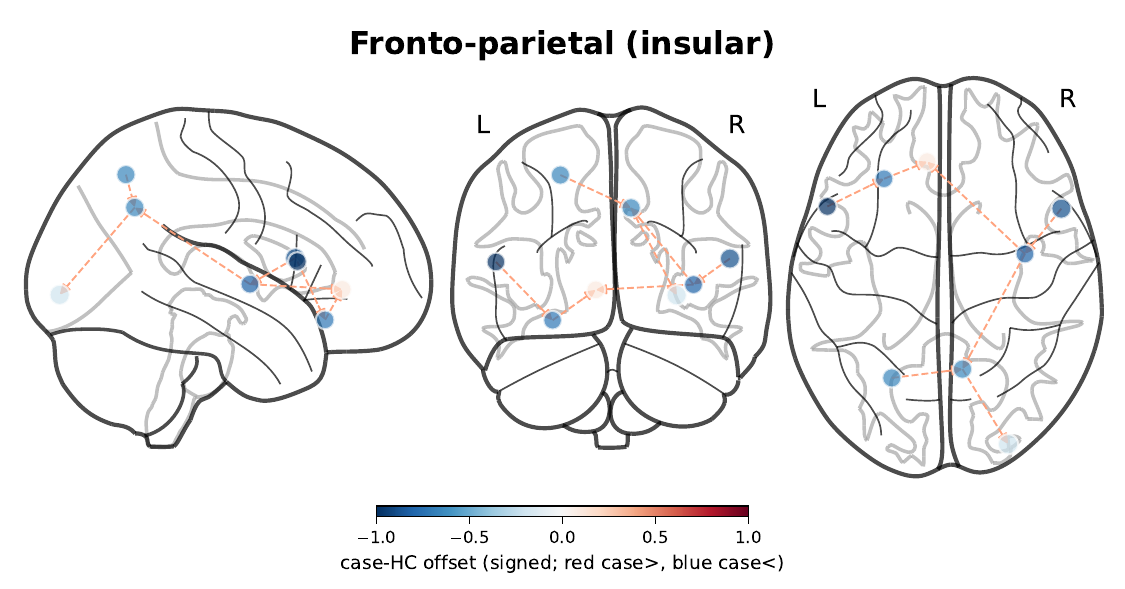}
\caption{FEP (fronto-parietal, insular).}
\label{fig:glassbrain-fep}
\end{subfigure}

\begin{subfigure}{0.62\textwidth}
\includegraphics[width=\textwidth]{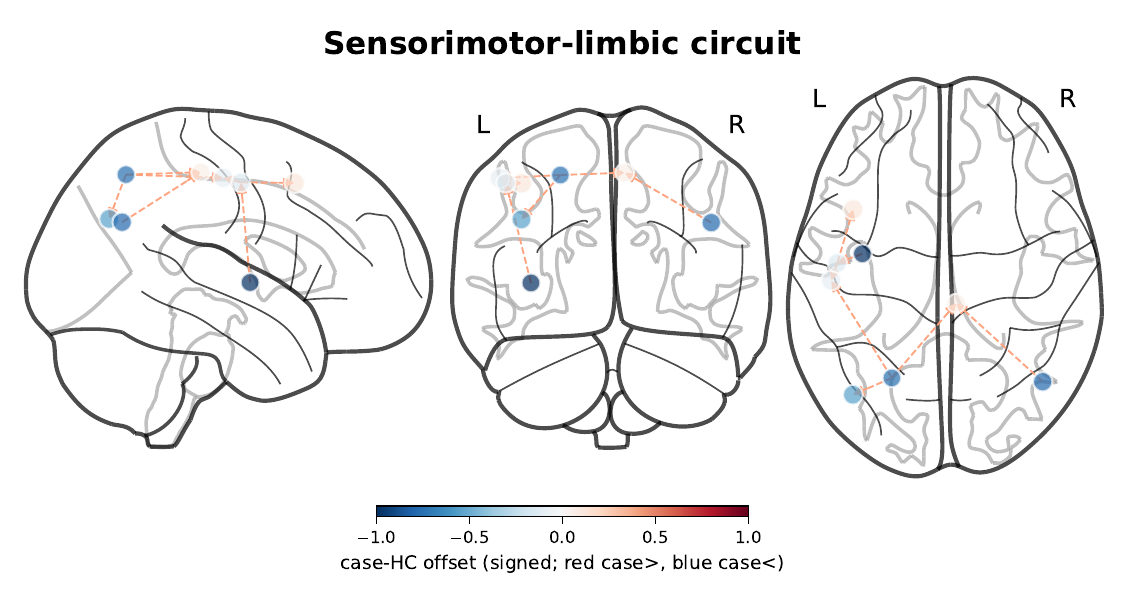}
\caption{PD (sensorimotor, insular).}
\label{fig:glassbrain-pd}
\end{subfigure}
\caption{\textbf{Disease pathway glass brains} (MNI ortho views). Two quantities are overlaid and should be kept distinct. \emph{Node selection and size} encode decision sensitivity: the eight ROIs with the highest unsigned amplitude attribution $|\partial\,\mathrm{logit}/\partial h_{\mathrm{health}}|$ (3-seed full-cohort, cross-seed stable). \emph{Node color} encodes the case$-$HC state offset $E[h\mid\mathrm{case}]-E[h\mid\mathrm{HC}]$ (red $=$ higher in patients, blue $=$ lower), per disease normalized; a group-level representational difference, separate from a decision quantity. The two can differ. Arrows are an illustrative minimum-spanning-tree pathway directed from smaller to larger offset, independent of learned connectivity. Circuit labels: MDD fronto-limbic-temporal, FEP fronto-parietal (insular), PD sensorimotor (insular). Colors and sizes are per-disease normalized, so absolute magnitudes are comparable only within a disease; offsets are group-level and correlational, without a causal claim, and deep ROIs (amygdala, hippocampus) are sLORETA source estimates.}
\label{fig:glassbrain}
\end{figure*}

\end{document}